\documentclass[preprint,12pt]{elsarticle}

\usepackage{amssymb}
\usepackage{amsmath}
\graphicspath{{./figures/}}
\usepackage{xcolor}
\definecolor{linkblue}{RGB}{0,80,160}
\usepackage{float}
\usepackage{hyperref}
\hypersetup{
    colorlinks=true,
    linkcolor=linkblue,
    citecolor=linkblue,
    urlcolor=linkblue,
    pdfborder={0 0 0}
}
\usepackage{verbatim}
\usepackage{graphicx}
\usepackage{subcaption}
\usepackage{amssymb}
\usepackage{multirow}
\usepackage{booktabs}
\usepackage{makecell}
\usepackage{adjustbox}
\journal{Engineering Applications of Artificial Intelligence}

\begin{document}

\begin{frontmatter}

\title{Structure-Guided Mixed Masked Pretraining and Spatial Continuity Regularization for Printed Circuit Board Defect Detection}

\author[label1]{Peitong Wang}
\ead{wd2334129@stu.ahu.edu.cn}

\author[label2]{Nuo Wang}
\ead{2575994185@qq.com}

\author[label1]{Enxin Qin}
\ead{19719272939@163.com}

\author[label1]{Chengjin Yu\corref{cor1}}
\ead{chengjinyu@ahu.edu.cn}

\author[label1,label3,label4]{Hanyu Xuan}
\ead{22176@ahu.edu.cn}

\author[label2]{Yuanting Yan}
\ead{ytyan@ahu.edu.cn}

\cortext[cor1]{Corresponding author.}

\affiliation[label1]{
    organization={School of Big Data and Statistics, Anhui University},
    city={Hefei},
    postcode={230601},
    country={China}
}

\affiliation[label2]{
    organization={School of Computer Science and Technology, Anhui University},
    city={Hefei},
    postcode={230601},
    country={China}
}

\affiliation[label3]{
    organization={School of Artificial Intelligence and Data Science, University of Science and Technology of China},
    city={Hefei},
    postcode={230026},
    country={China}
}

\affiliation[label4]{
    organization={Institute of Dataspace, Hefei Comprehensive National Science Center},
    city={Hefei},
    postcode={230088},
    country={China}
}

% -----------------------------
% Abstract + Keywords
% -----------------------------

% The main difficulties arise from weak defect appearance, dense circuit backgrounds, insufficient structural context modeling, and fragmented localization on thin continuous regions.
\begin{abstract}
Printed circuit board (PCB) defect detection is an essential part of automated optical inspection (AOI); yet it remains challenging in practice because many defects are tiny, low-contrast, and embedded in dense circuit backgrounds. To address these issues, this paper presents a two-phase PCB defect detection framework that combines structure-guided mixed masked pretraining with spatial continuity regularization. 
In the pretraining stage, we design a sparse convolutional masked pretraining scheme to exploit unlabeled PCB images, where structure-guided mixed masking is used to construct informative masked inputs. 
The sparse convolutional reconstruction pipeline suppresses invalid responses from masked regions and enables the detector backbone to infer missing PCB structures from visible conductive patterns, thereby learning PCB structural priors. 
In the fine-tuning stage, the pretrained backbone is transferred to the downstream defect detection task. For the task, a spatial continuity regularization term is introduced during fine-tuning. 
This term constrains dispersed positive predictions assigned to the same defect instance and promotes more compact localization on elongated defect regions. Experiments on the DsPCBSD+ dataset show that the proposed method achieves 85.5\% $\mathrm{mAP}_{0.5}$ and 52.3\% $\mathrm{mAP}_{0.5:0.95}$, outperforming several strong baseline detectors. 
Ablation studies and qualitative results further confirm the effectiveness of the proposed framework for robust PCB defect detection in industrial AOI scenarios.
\end{abstract}

\begin{keyword}
Printed circuit board defect detection \sep YOLOv8 \sep masked image modeling \sep self-supervised pretraining \sep spatial continuity regularization \sep automated optical inspection
\end{keyword}

\end{frontmatter}

% =========================================================
\section{Introduction}
\label{sec:introduction}
Printed circuit boards (PCBs) are fundamental components in modern electronic devices, providing electrical connections among different functional units~\cite{tang2024lightweight,angelopoulos2019tackling}. 
During PCB manufacturing, defects may be introduced in several processes, such as pressing, drilling, copper deposition, and application of dry film~\cite{TAN2020121892}. Reliable PCB defect detection is essential for quality control in electronic manufacturing. 
%In this work, we focus on nine common defect types in industrial production: short, spur, spurious copper, open circuit, mouse bite, missing hole (hole breakout), conductor scratch, foreign conductor object, and foreign base material foreign object~\cite{lv2024dataset}. Fig.~\ref{fig:problem}(a) summarizes the AOI inspection setting and shows representative samples of these defect categories.
% 印刷电路板是现代电子设备的基础组件，为不同功能单元之间提供电气连接。在生产过程中，多个工序均可能引入缺陷。因此，可靠的PCB缺陷检测对电子制造的质量控制至关重要。本文聚焦工业生产中常见的九类缺陷：短路、毛刺、多余铜、开路、缺口、漏孔、导线划伤、异质导电异物和基材异物。图示展示了自动光学检测场景及各类缺陷的典型示例。

Traditional PCB defect detection mainly relies on manual visual inspection. This process is highly dependent on the inspector's experience and is often affected by subjectivity, high labor costs, and limited consistency~\cite{MOGANTI1996287}. As electronic products move toward higher density and greater functional integration, PCB layouts have become more complex, while defects are often smaller and harder to distinguish. 
Manual inspection is being replaced by automated optical inspection (AOI), which has been widely used in industrial production to improve detection accuracy and efficiency~\cite{ZHOU2023557,10044670}. 
More recently, deep learning (DL) has further promoted the development of AOI algorithms. Compared with traditional pipelines based on image processing and machine learning, deep learning methods can provide more robust defect detection while maintaining high inspection speed~\cite{KANG2023120121}.
% 传统的PCB表面缺陷检测主要依赖人工目检。这种方式高度依赖检测人员的经验，且易受主观性、高人力成本和一致性差等问题影响。由于电子产品正朝着高密度和高功能集成的方向发展，PCB设计日趋复杂，而缺陷则变得更微小、更难以识别。因此，人工检测正逐步被自动光学检测所取代，后者因能显著提升检测的准确性和效率，已在工业生产中得到广泛应用。近年来，深度学习进一步推动了端到端AOI算法的发展。与传统的图像处理和机器学习流程相比，基于深度学习的方法能够在保持高速检测的同时，提供更稳健的缺陷识别能力。

\begin{figure}[!htb]
  \centering
  \includegraphics[width=1.0\textwidth]{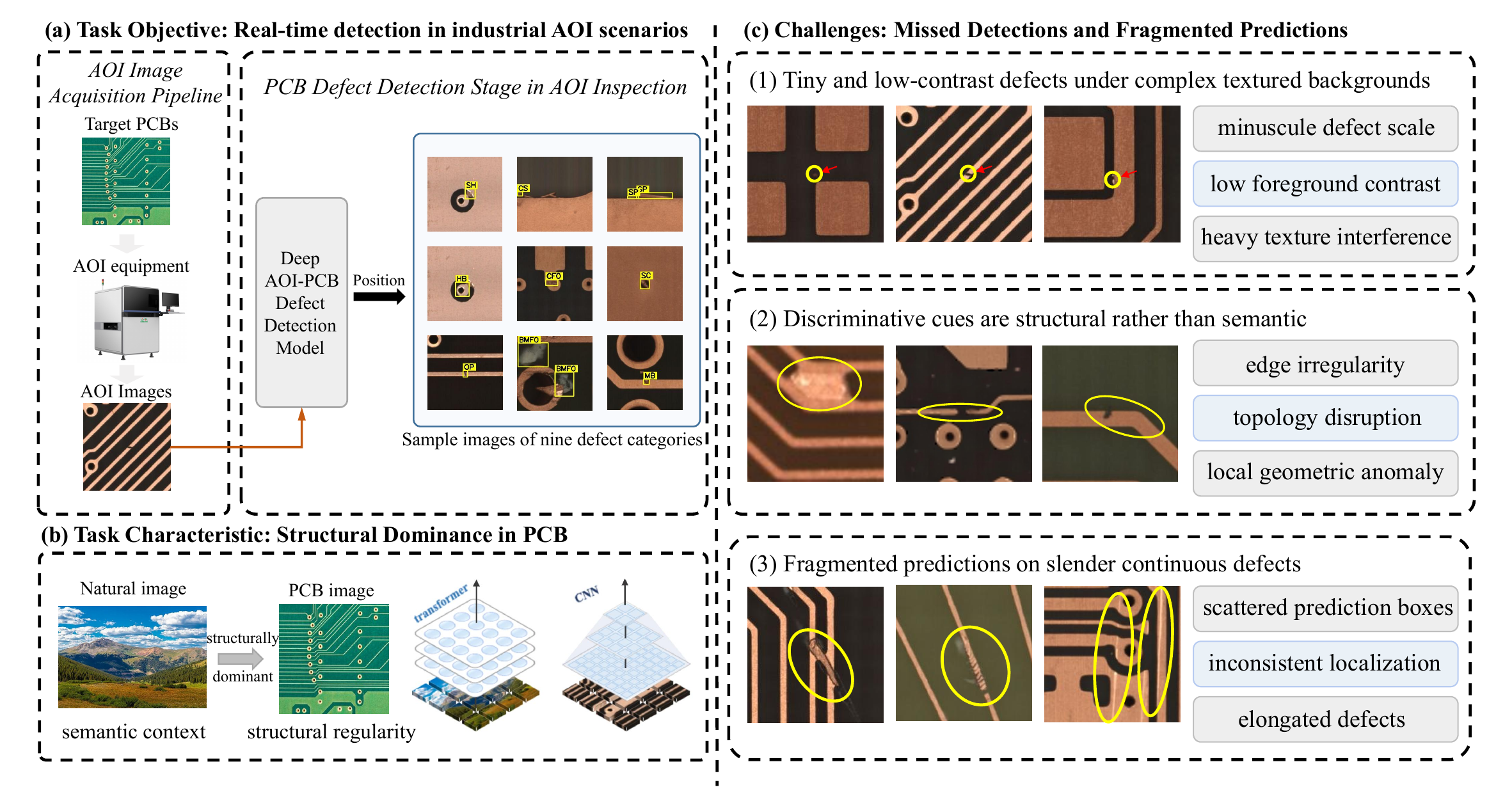}
  \caption{Overview of the motivation and key observations for industrial AOI PCB defect detection. (a) The AOI inspection pipeline is the mainstream solution for modern PCB defect detection. (b) PCB defect inspection is more suitable for CNN-based detectors than Transformer-based models. (c) Three typical difficulties in PCB defect detection are illustrated.}
  \label{fig:problem}
\end{figure}

Deep learning methods for PCB defect inspection can be broadly divided into classification, detection, and segmentation approaches~\cite{SUN2025128101}. Among them, object detection is the most widely adopted setting for industrial AOI. And it is commonly categorized into two-stage and one-stage frameworks~\cite{MENG2025113578}. Two-stage detectors, represented by Faster R-CNN~\cite{7485869}, first generate region proposals and then refine classification and localization, whereas one-stage detectors such as YOLO~\cite{10129965}, SSD~\cite{liu2016ssd}, and CenterNet~\cite{Duan_2019_ICCV} directly predict categories and locations in a single pass, making them more attractive for real-time deployment. In parallel, many vision models based on Transformers have shown strong ability in modeling global context and dependencies. Representative examples include ViT~\cite{dosovitskiy2020image}, Swin-Transformer~\cite{Liu_2021_ICCV}, DETR~\cite{carion2020end}, and Co-DETR~\cite{zong2023codetr}. 

%However, PCB defect detection has different visual characteristics from natural image understanding. As shown in Fig.~\ref{fig:problem}(b), it depends less on rich semantic context. Its discriminative cues are more often reflected by structural regularity and local geometric anomalies. These cues include edge irregularities, topology disruptions, and variations in line width. This task characteristic makes a CNN-based detector more suitable, since convolution is naturally aligned with local structure modeling and hierarchical feature extraction, while also providing higher efficiency and easier deployment in industrial AOI scenarios. Therefore, we adopt YOLOv8 as the baseline detector in this work.

% 用于PCB缺陷检测的深度学习方法大致可分为基于分类、基于检测和基于分割的方法。其中，目标检测是工业自动光学检测中最常采用的设置，通常又可分为两阶段与一阶段框架。以Faster R-CNN为代表的两阶段检测器先生成候选区域，再进行分类与定位优化；而以YOLO、SSD和CenterNet为代表的一阶段检测器在单次前向传播中直接完成类别与位置预测，因此在实时部署中更具吸引力。

% 与此同时，基于Transformer的视觉模型，如ViT、Swin Transformer、DETR和Co-DETR，在捕捉长程依赖关系和全局上下文方面展现出较强能力。然而，如图所示，PCB缺陷检测相比自然图像理解，并不那么依赖丰富的语义上下文，其判别线索更多体现为结构规则性和局部几何异常，例如边缘不规则、拓扑破坏和线宽变化等。这一任务特性使得基于卷积神经网络的检测器更为适合，因为卷积操作天然契合局部结构建模与层级特征提取，同时在工业自动光学检测场景中具备更高的效率和更好的部署实用性。因此，本文选用YOLOv8作为基础检测器。

Although deep learning has improved PCB defect detection, practical inspection still faces several prominent challenges. As illustrated in Fig.~\ref{fig:problem}(c), these challenges can be understood progressively from low-level appearance, to mid-level structural representation, and finally to output-level localization consistency: \textbf{(1) At the visual level, PCB defects are often tiny, subtle, and sparsely distributed~\cite{ma2024surface}.} The surrounding regions usually contain dense circuit layouts and highly textured backgrounds. This makes separation between foreground and background difficult and weakens feature discriminability~\cite{zhu2025novel}. \textbf{(2) At the  representation level, the main challenge is how to capture structural context.} Transformer-based models can capture long-range dependencies and global contextual relationships, whereas CNN-based detectors mainly rely on local convolution and hierarchical feature aggregation~\cite{khan2023survey}. For defects that require stronger semantic or structural association across a wider region, CNN-based models may fail to capture sufficient contextual cues. This limitation makes these defects difficult to detect.  \textbf{(3) At the output level, one-stage detectors tend to produce fragmented predictions for slender and spatially continuous defects, splitting a single defect region into multiple scattered boxes~\cite{rs16162910}}. Although post-processing methods can heuristically merge overlapping boxes, they do not explicitly enforce spatial coherence during representation learning and thus cannot fundamentally resolve this issue. In addition to these visual and localization challenges, limited labeled data is another factor that restricts the practical performance of PCB defect detectors~\cite{app14156774}.
% 尽管深度学习已显著提升了PCB缺陷检测水平，但在实际工业检测中仍面临诸多突出挑战。如图1所示，这些挑战可由浅入深地分为低层外观、中层结构表征以及输出层定位一致性三个层面。在低层视觉层面，PCB缺陷通常尺度微小、对比度低且分布稀疏，而其周围图像往往包含密集的线路布局和强纹理背景，导致严重的前景‑背景混淆和较弱的特征可判别性。该问题在光照变化、成像噪声以及部分缺陷类别间高度视觉相似性的影响下会进一步加剧。在中层表征层面，主要挑战在于结构上下文的建模。基于Transformer的模型能够捕捉长程依赖和全局上下文关系，而基于卷积神经网络的检测器则主要依赖局部卷积和层级特征聚合。对于需要在更广区域建立更强语义或结构关联的缺陷，基于卷积神经网络的方法可能难以捕获足够的上下文线索，导致这类缺陷难以被有效检测。在输出层面，一阶段检测器在面对细长且空间连续的缺陷时，容易产生碎片化预测，即将同一缺陷区域分裂为多个分散的检测框。虽然后处理方法能够启发式地合并重叠框，但它们未在表征学习阶段显式强化空间一致性，因而无法从根本上解决这一问题。

To address these challenges, we propose a structure-guided mixed masked pretraining framework and a spatial continuity regularization for
PCB defect detection. 
In the pretraining stage, we introduce SG-MIM, a sparse convolutional masked pretraining scheme that adapts masked image modeling to the YOLOv8s backbone using unlabeled PCB images.
A structure-guided mixed masking strategy is used to construct informative masked inputs by incorporating PCB structural cues. 
Different from standard dense convolution, the proposed sparse convolutional reconstruction pipeline propagates visibility information through the convolutional backbone and suppresses invalid responses from masked regions. 
In fine-tuning stage, the pretrained YOLOv8s backbone is transferred to supervised defect detection. 
A spatial continuity regularization loss is then added during fine-tuning to constrain the distribution of positive predictions. 
It penalizes overly scattered responses and encourages more compact localization for slender and continuous defects. 

% 为在保持基于卷积神经网络检测器高效性的同时应对上述挑战，我们在YOLOv8基础上构建了一个两阶段学习策略。在阶段A，我们引入了结构引导掩码图像建模方法，用于在无标注PCB图像上对YOLO骨干网络进行预训练。SG-MIM并非完全依赖随机掩码，而是采用一种混合掩码策略，将PCB结构线索（如边缘、走线和局部几何变化）融入掩码过程中。这促使骨干网络从缺陷敏感区域中学习，增强其对微弱、低对比度结构的表征能力。为使掩码预训练适配卷积骨干网络并提升其上下文建模能力，我们在轻量级重建流程中进一步引入了稀疏卷积。该设计可减少被掩码区域的干扰，并帮助YOLO骨干网络从可见部分推断缺失的PCB结构。

% 在阶段B，将预训练后的YOLOv8骨干网络迁移至有监督缺陷检测任务。随后在微调过程中加入空间连续性正则化损失，以约束同类正样本预测的分布。该损失通过惩罚过度分散的响应，促使模型对细长连续缺陷产生更紧凑的定位结果。

Our main contributions are summarized as follows:
\begin{enumerate}
  \item We propose a structure-guided mixed masking strategy for model pretraining. By leveraging PCB structural cues to guide the masking process on unlabeled images, the proposed strategy encourages the backbone to focus on defect-sensitive regions and improves the representation of tiny, low-contrast, and structure-sensitive defects.
% 我们提出了一种结构引导掩码策略，用于 PCB 缺陷预训练。该策略利用 PCB 图像中的结构线索来引导无标注图像上的掩码生成过程，从而促使骨干网络更加关注对缺陷敏感的区域，并提升对微小、低对比度及结构敏感缺陷的表征能力。

  \item We incorporate sparse convolution into the YOLOv8s backbone pretraining. This design adapts masked pretraining to the convolutional backbone by reducing the interference from masked regions, while helping the model infer missing PCB structures from visible contextual cues. As a result, the backbone can learn more transferable structural representations for downstream defect detection.
%我们通过类似U-Net的重建流程，将稀疏卷积引入YOLOv8骨干网络的预训练范式中。该设计使掩码预训练适配卷积骨干网络，不仅能减少掩蔽区域的干扰，同时帮助模型从可见的上下文线索中推断缺失的PCB结构。最终，骨干网络能够学习更具可迁移性的结构表征，以提升下游缺陷检测的性能。

  \item We introduce a spatial continuity regularization loss to alleviate fragmented predictions on slender and spatially continuous defects. By penalizing overly dispersed same-defect predictions, the proposed loss encourages more compact and coherent localization results.
% 我们引入了一种空间连续性正则化损失，以缓解细长且空间连续缺陷上的碎片化预测问题。该损失通过惩罚同一缺陷预测的过度离散分布，促使模型输出更加紧凑、连贯的定位结果。

  \item We conduct extensive experiments and ablation studies to validate the proposed framework. The results show consistent improvements in PCB defect detection and more coherent localization compared with strong baselines.
\end{enumerate}

% =========================================================
\section{Related Work}
\label{sec:related}
\subsection{YOLO-based Methods for PCB Defect Detection}
YOLO detectors adopt a modular and scalable architecture~\cite{sohan2024review}. 
This design supports flexible feature fusion and multi-level detection heads for robust multi-scale localization~\cite{yaseen2024yolov8indepthexplorationinternal}. These characteristics make YOLO a popular choice for industrial surface inspection, including PCB defect detection.

Recent studies have improved YOLO-style frameworks from multiple perspectives. For lightweight deployment and real-time inference, YOLO-CEA ~\cite{Zhao2024} reduces computational cost by optimizing MAdds, thereby achieving fast inference with robust detection performance. Similarly, MobileNet-YOLO-Fast ~\cite{10.1117/1.JEI.30.4.043004} was introduced for rapid PCB defect inspection with a compact model size. To enhance tiny-defect perception under complex PCB backgrounds, several works incorporate attention mechanisms into the backbone. For example, PCB-YOLO ~\cite{tang2023pcb} introduces a Swin-Transformer and a joint attention mechanism to suppress background interference, while a YOLOX-based lightweight detector with coordinated attention further strengthens the recognition of minor defects on PCB surfaces ~\cite{9905499}. Beyond architectural enhancements, other efforts focus on improving localization accuracy by redesigning bounding-box regression losses, such as modified IoU-based objectives to refine box regression output within YOLOv4-style detectors ~\cite{LIU2022116178}. More recently, YOLO-HMC has integrated stronger feature extractors, attention-based defect highlighting, and context-aware upsampling to better capture tiny defects and aggregate semantic information under complex backgrounds ~\cite{yuan2024yolo}.

Despite these advances, the training of YOLO-based PCB detectors still relies on limited labeled information. This paradigm does not fully exploit the rich knowledge embedded within the images~\cite{ling2023printed,zhou2023review}. Since YOLO backbones are hierarchical and convolutional, generative self-supervised pretraining provides a feasible way to exploit abundant unlabeled PCB images. By learning PCB structural priors before fine-tuning, the backbone can obtain more discriminative representations for downstream defect detection.
% YOLO检测器采用模块化、可扩展的网络结构，使其能够灵活定制特征融合设计与多层检测头，从而实现更鲁棒的多尺度定位。这些特性使YOLO在工业表面检测（包括PCB缺陷检测）中被广泛采用。
% 近期研究从多个角度改进了YOLO类框架。在轻量化部署与实时推理方面，YOLO-CEA \cite{Zhao2024} 通过优化 Madds 降低计算开销，从而在保证鲁棒检测性能的同时实现快速推理。类似地，MobileNet-YOLO-Fast \cite{10.1117/1.JEI.30.4.043004} 也被提出用于紧凑模型规模下的快速PCB缺陷检测。为提升复杂PCB背景下对微小缺陷的感知能力，不少工作在骨干网络或颈部结构中引入注意力机制。例如，PCB-YOLO \cite{tang2023pcb} 引入 Swin Transformer 与联合注意力机制以抑制背景干扰；而基于YOLOX的轻量化检测器则结合坐标注意力进一步强化PCB表面微小缺陷识别能力 \cite{9905499}。除结构改进外，还有研究通过重新设计边界框回归损失以提升定位精度，例如在YOLOv4类检测器中采用改进的IoU目标来细化回归输出 \cite{LIU2022116178}。此外，YOLO-HMC通过融合更强的特征提取器、注意力缺陷增强模块与上下文感知上采样结构，更好地捕获微小缺陷并在复杂背景下聚合语义信息 \cite{yuan2024yolo}。
% 尽管已有这些进展，基于YOLO的PCB缺陷检测器仍受制于标注数据的有限性，这可能影响所学习特征表示的质量。现有方法大多直接使用在通用数据集上预训练的权重，或以完全监督的方式训练检测器，而针对PCB缺陷检测领域的专门预训练仍探索不足。由于YOLO骨干网络具有层次化与卷积的特性，生成式自监督预训练为利用大量无标注PCB图像提供了一条可行的途径。通过在微调前学习PCB特有的结构先验，骨干网络能够获得更具判别力的特征表示，从而提升下游缺陷检测的性能。

\subsection{Self-Supervised Pretraining and Masked Autoencoders}

In industrial AOI scenarios, obtaining pixel-level or instance-level annotations at scale is costly and time-consuming~\cite{tao2022deep}. Consequently, deep models often suffer from limited training data and weaker generalization. Self-supervised learning (SSL) alleviates this bottleneck by exploiting abundant unlabeled images through pretext objectives to learn transferable visual representations for downstream tasks ~\cite{jing2020self}.

Earlier visual SSL research was mainly dominated by contrastive learning, which formulates self-supervised learning as an instance discrimination task~\cite{oord2018representation,he2020momentum,chen2020simple} . To address issues such as mode collapse and to improve representation quality, a series of non-contrastive and clustering-based variants were further developed~\cite{grill2020bootstrap,caron2020unsupervised,chen2021exploring} . With the rise of vision transformers ~\cite{vaswani2017attention}, however, masking-based generative pretraining gradually became a major direction in visual self-supervised learning.

Inspired by masked language modeling in natural language processing ~\cite{devlin2019bert}, masked image modeling (MIM) has emerged as an effective visual pretraining paradigm. Early works explored predicting discrete token indices of masked patches~\cite{bao2021beit}, while representative methods such as MAE~\cite{he2022masked} and SimMIM~\cite{xie2022simmim} directly regress raw RGB values of masked regions, greatly simplifying the pretraining objective. Subsequent studies further validated the effectiveness of masking-based pretraining on vision transformers ~\cite{zhou2023self}.

Nevertheless, most masked autoencoder methods were originally developed for token-based Transformer architectures~\cite{hondru2025masked}. This makes them less compatible with hierarchical convolutional backbones, which are still widely used in practical detection frameworks. SparK~\cite{tian2023designing} has shown that masking-based pretraining can be extended to CNNs through sparse convolution and hierarchical reconstruction. Motivated by this idea, we further investigate masked pretraining for convolutional AOI defect detection by developing a PCB-oriented strategy for the YOLOv8 backbone. In this way, large-scale unlabeled PCB images can be directly exploited to learn domain-specific structural priors, which are beneficial for downstream defect detection.

\section{Methodology}
\subsection{Overall Framework}
\begin{figure}[!htb]
  \centering
  \includegraphics[width=1.0\textwidth]{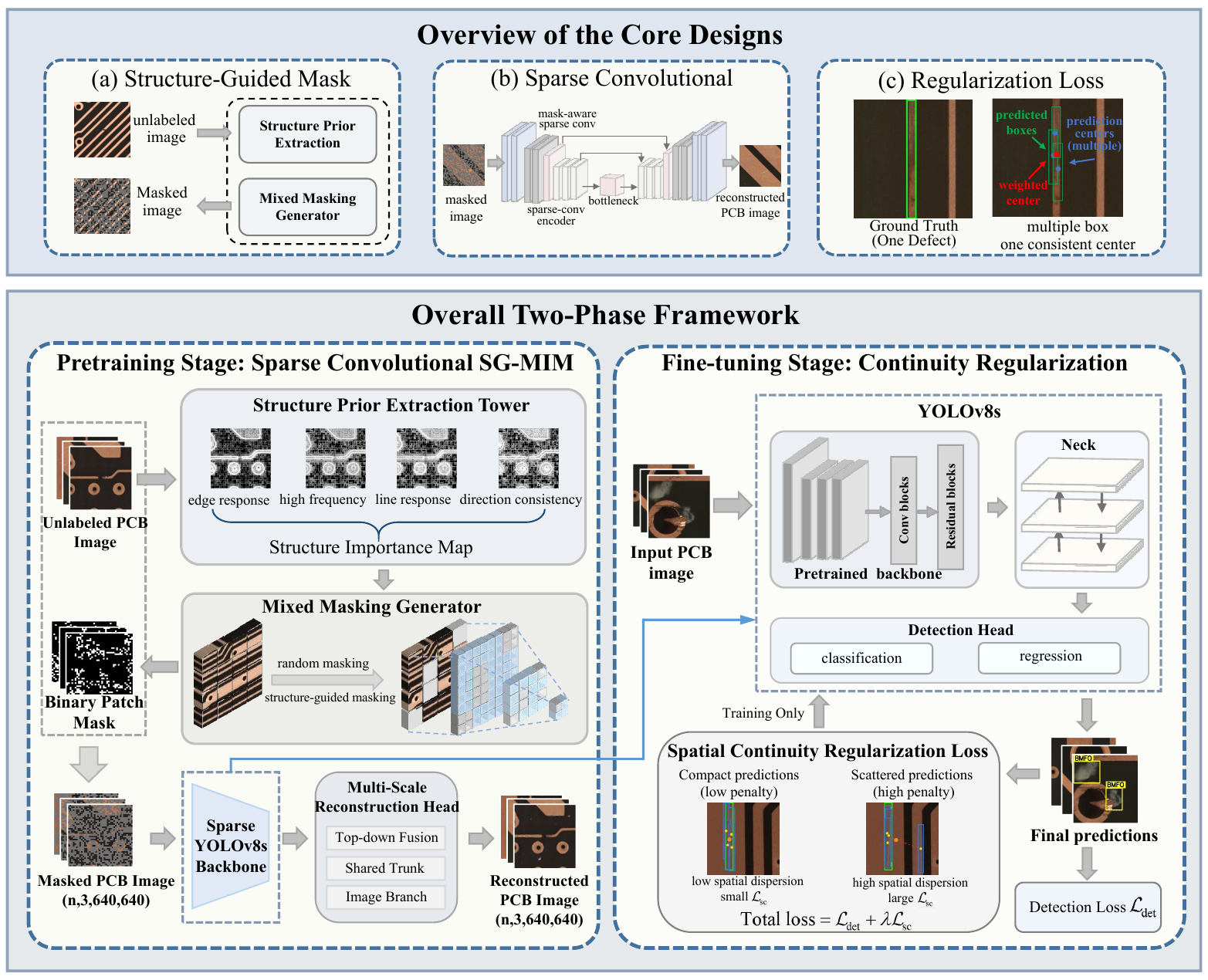}
\caption{Overall framework and core designs of the proposed method.
(a) Structure-guided mixed masking generates informative masked PCB inputs using structural cues. 
(b) Sparse convolution suppresses invalid responses in the masked region. 
(c) Spatial continuity regularization constrains scattered predictions assigned to the same defect instance. 
In the pretraining stage, the backbone is pretrained on unlabeled PCB images; in the fine-tuning stage, the pretrained backbone is transferred to YOLOv8s, where the continuity regularization loss improves localization coherence without adding inference cost.}
  \label{fig:framework}
\end{figure}

As illustrated in Fig.~\ref{fig:framework}, the proposed method consists of three core designs: (a) structure-guided mixed masking, (b) sparse convolutional, and (c) spatial continuity regularization. The lower part demonstrates how these components are integrated into one unified training pipeline.

In Stage-A, Structure-Guided Masked Image Modeling (SG-MIM) is used for generative pretraining on unlabeled PCB images. A structure prior extraction module first estimates several structural cues, including edge response, high-frequency response, line response, and direction consistency. These cues are combined to produce a structure importance map. Based on this map, a mixed masking generator selects either random masking or structure-guided masking to produce masked PCB inputs. The masked images are then fed into a sparse YOLOv8 backbone encoder. A multi-scale reconstruction head is used to recover PCB structures. Through this pretraining task, the backbone learns PCB-specific structural priors and captures contextual associations among local PCB patterns from unlabeled data.

In Stage-B, the pretrained backbone is transferred to the YOLOv8 detector for supervised fine-tuning on labeled PCB images. To reduce fragmented predictions on slender continuous defects, we introduce a spatial continuity regularization loss during training. This loss encourages predictions from the same defect to remain spatially compact and coherent. It also reduces redundant and scattered boxes. 
The regularization term is used only during training and removed during inference. Thus, the proposed framework improves structural representation and localization consistency without adding extra test-time cost.
% 如图1所示，所提出的方法包含三个核心设计以及一个用于PCB缺陷检测的整体两阶段训练框架。上半部分展示了三个关键组成部分：结构引导掩码、稀疏卷积重建和空间连续性正则化。下半部分则说明了这些组件如何被集成到基于YOLOv8的训练流程中。

% 在阶段A，我们采用结构引导掩码图像建模对无标注PCB图像进行生成式预训练。结构先验提取模块首先估计多种结构线索，包括边缘响应、高频响应、线形响应和方向一致性。这些线索被组合生成一个结构重要性图。基于此图，一个混合掩码生成器选择随机掩码或结构引导掩码来生成掩码后的PCB输入。掩码图像随后被输入稀疏卷积YOLOv8骨干编码器，并通过多尺度重建头恢复有意义的PCB结构。通过此预训练任务，骨干网络能够从未标注数据中学习PCB特有的结构先验，并捕捉局部PCB模式间的上下文关联。

% 在阶段B，预训练后的骨干网络被迁移至YOLOv8检测器，在有标注的PCB图像上进行有监督微调。为减少对细长连续缺陷的碎片化预测，我们在训练过程中引入了空间连续性正则化损失。该损失促使同一缺陷的预测在空间上保持紧凑和连贯，同时减少冗余和分散的检测框。正则化项仅在训练时使用，在推理阶段被移除。因此，所提框架能够在不增加额外推理开销的情况下，提升结构表征能力和定位一致性

\subsection{Stage-A: Generative Pretraining via SG-MIM}
% \subsection{Stage-A：基于结构引导掩码图像建模的生成式预训练}
\begin{figure}[t]
  \centering
  \includegraphics[width=1.0\textwidth]{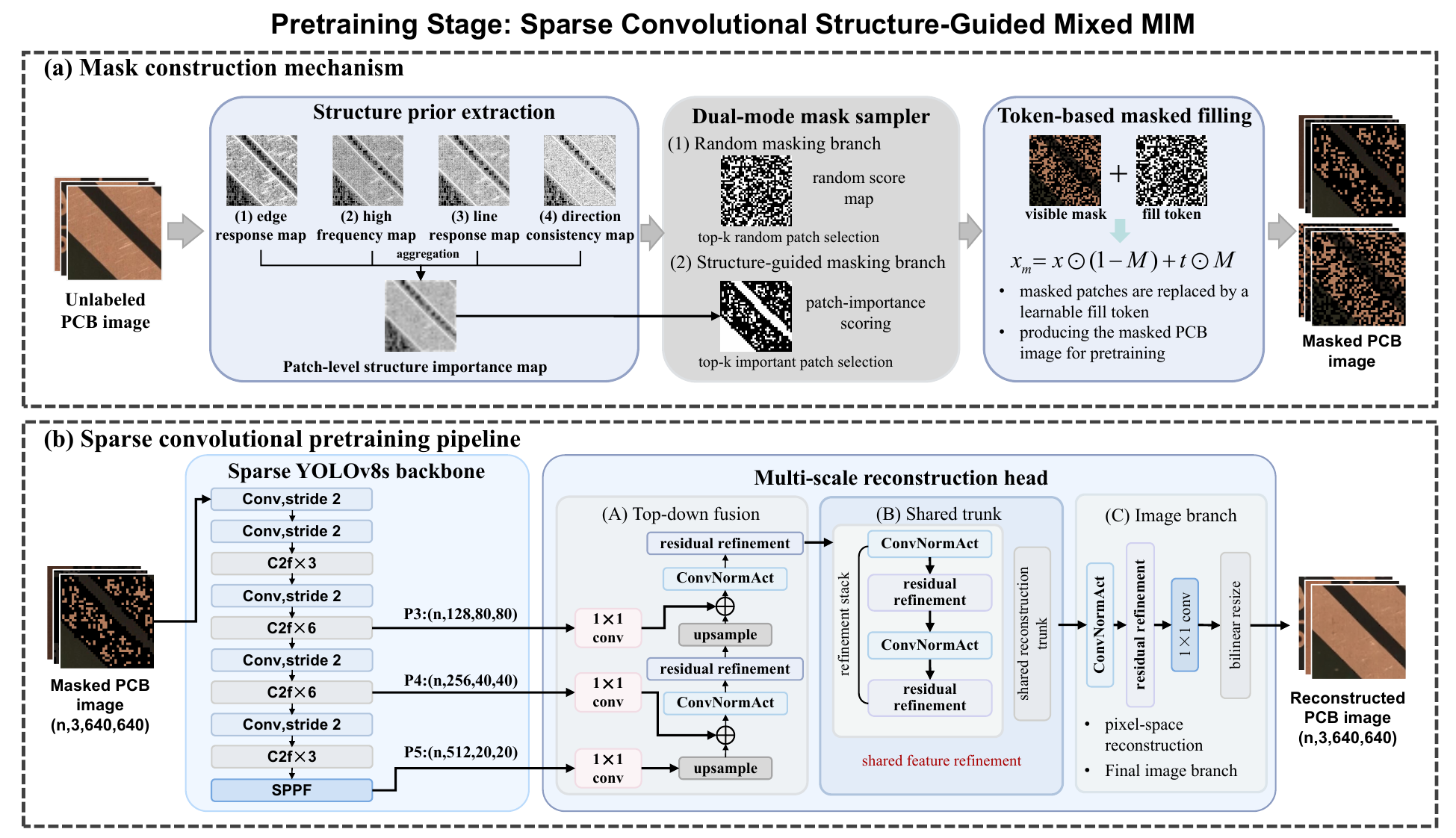}
  \caption{Overview of the pretraining stage with sparse-convolutional structure-guided mixed masked image modeling. (a) Structure-guided mixed masking constructs informative masked PCB inputs by combining structural cues with random masking. (b) The masked image is reconstructed through a sparse YOLOv8s backbone and a lightweight multi-scale reconstruction head, enabling the model to suppress invalid responses in the masked region and learn PCB structural priors from visible context.}
  \label{fig:StageA}
\end{figure}

In Stage-A, we use a structure-guided mixed masked generative task to pretrain the detector backbone on unlabeled PCB images. As shown in Fig.~\ref{fig:StageA}, this stage contains three main steps: (1) structure prior extraction with mixed masking, (2) sparse encoding with mask guidance, and (3) multi scale reconstruction. 
% PCB images usually contain dense traces, repeated layouts, and fine geometric structures. For this reason, the masking process is not purely random. Instead, it introduces PCB structural cues to guide part of the mask generation. The pretraining task asks the backbone to recover missing PCB patterns from visible regions. It also helps the model learn PCB specific structural priors and contextual relations among local PCB patterns before downstream fine-tuning.
% 阶段A通过一项结构感知的掩码生成任务，在无标注PCB图像上对检测器骨干网络进行预训练。如图2所示，该流程包含三个主要步骤：基于结构先验的混合掩码生成、稀疏卷积骨干编码，以及仅针对掩码区域的多尺度重建优化。PCB图像包含密集的走线、重复的布局和精细的几何结构，因此掩码过程由PCB特有的结构线索引导，而非完全随机。该预训练任务促使骨干网络从可见区域中恢复缺失的PCB图案，并在下游微调前学习可迁移的结构表征。

\subsubsection{Structure Prior Extraction with Mixed Masking}

Unlike natural images, PCB images concentrate most discriminative information on conductive traces, pad boundaries, and local routing patterns. In practice, many defects are small and have weak visual responses. Meanwhile, the background often contains textured regions that are less relevant to defect recognition. Under this condition, purely random masking may allocate many masked patches to visually unimportant areas. 
% 与自然图像不同，PCB图像的大部分判别信息集中在导电走线、焊盘边界和局部布线图案上。在实际场景中，许多缺陷尺寸微小且视觉响应微弱，而背景常包含与缺陷识别无关的纹理区域。在此情况下，完全随机的掩码可能将许多图像块分配到视觉不重要的区域，这会削弱重建目标的有效性，并降低模型对关键结构区域的关注压力。为使掩码预训练与PCB检测任务更契合，我们首先从每张无标注图像中构建结构先验，如图2（a）所示

%The high-frequency cue combines the Laplacian response and local variance. It captures local detail strength and abrupt structural changes.

To align masked pretraining with PCB inspection, we first construct a structure prior from each unlabeled image, as shown in Fig.~\ref{fig:StageA}(a). 
Specifically, given an unlabeled PCB image $\mathbf{x}\in\mathbb{R}^{3\times H\times W}$, we first convert it into a grayscale image $\mathbf{I}$ and extract four complementary structural cues.

(1) The edge response map $\mathbf{E}$ is computed from the image gradient magnitude to highlight trace boundaries and pad edges~\cite{canny2009computational}:
\begin{equation}
\mathbf{E}=\sqrt{(\partial_x \mathbf{I})^2+(\partial_y \mathbf{I})^2}.
\end{equation}

(2) The high-frequency map $\mathbf{H}$ combines the Laplacian response and local variance to capture fine details and abrupt structural changes~\cite{marr1980theory,haralick2007textural}:
\begin{equation}
\mathbf{H}=\alpha |\Delta \mathbf{I}|+(1-\alpha)\operatorname{Var}_{\mathcal{N}}(\mathbf{I}).
\end{equation}

(3) The line response map $\mathbf{L}$ is obtained by applying multi-scale and multi-orientation line-sensitive filters to enhance conductive traces and elongated structures~\cite{freeman1991design}:
\begin{equation}
\mathbf{L}=\max_{\sigma\in\mathcal{S},\,\theta\in\Theta}
\left| \mathbf{I} * \mathbf{K}^{line}_{\sigma,\theta} \right|.
\end{equation}
where $*$ denotes convolution, and $\mathcal{S}$ and $\Theta$ denote the sampled scales and orientations.

(4) The direction consistency map $\mathbf{D}$ is computed from the eigenvalues of the local structure tensor to measure local orientation consistency~\cite{bigun2002multidimensional}:
\begin{equation}
\mathbf{D}=\frac{\lambda_1-\lambda_2}{\lambda_1+\lambda_2+\epsilon},
\end{equation}
where $\lambda_1$ and $\lambda_2$ are the two eigenvalues of the structure tensor.

Finally, all response maps are normalized and projected into patch space by average pooling. 
For the $i$-th patch, its structure importance score is calculated as
\begin{equation}
s_i=w_E\bar{E}_i+w_H\bar{H}_i+w_L\bar{L}_i+w_D\bar{D}_i,
\end{equation}
where $\bar{E}_i$, $\bar{H}_i$, $\bar{L}_i$, and $\bar{D}_i$ denote the patch-level averaged responses, and $w_E$, $w_H$, $w_L$, and $w_D$ are weighting coefficients. 
The resulting patch-level score map is used as the structure importance map for subsequent structure-guided mask sampling.
% 具体而言，对于一张无标注 PCB 图像 $\mathbf{x}\in\mathbb{R}^{3\times H\times W}$，我们提取四类互补的结构线索，包括边缘响应图、高频组合响应图、线响应图以及方向一致性图。其中，高频线索由 Laplacian 响应与局部方差共同构成，用于联合表征局部细节强度与突变结构变化。上述响应图分别经过归一化后被映射到 patch 空间，并进一步线性聚合为 patch 级结构重要性图，用于衡量每个非重叠 patch 的结构显著性。

Based on the above maps, we use a dual-mode mask sampler consisting of a random masking mode and a structure-guided masking mode. It should be emphasized that the two modes are not fused simultaneously. Instead, one mode is sampled for each training image to generate the binary patch mask. The random mode preserves masking diversity by selecting masked patches according to random scores. The structure-guided mode performs top-$k$ selection according to patch importance scores. As a result, structurally salient regions are more likely to be masked. In this way, the model is encouraged to infer critical patterns and local structural details from the remaining visible context.
% 在此基础上，本文采用一个由随机掩码模式和结构引导掩码模式组成的双模式掩码采样器。需要强调的是，这两种模式并不会被同时融合，而是对每张训练图像按概率采样其中一种模式来生成二值 patch 掩码。随机模式通过随机得分进行遮挡选择，以保持掩码模式的多样性；结构引导模式则依据 patch 重要性得分进行 top-$k$ 选择，从而优先遮挡结构显著区域。借助这种设计，模型被迫根据剩余可见上下文去恢复关键导电模式以及与缺陷相关的局部结构。

Let $\mathbf{M}\in\{0,1\}^{N_p}$ denote the binary mask defined over $N_p$ image patches. 
Here, $\mathbf{M}_i=1$ means that the $i$-th patch is selected for masking, while the remaining patches are kept visible. 
For implementation, this patch mask is expanded to the original image resolution, resulting in $\mathbf{M}^{\uparrow}\in\{0,1\}^{1\times H\times W}$. 
The corresponding visible region is then represented by $\mathbf{V}=1-\mathbf{M}^{\uparrow}$.

A simple choice is to set the masked regions to zero. 
However, zero filling may introduce artificial blank areas that are not consistent with real PCB images. 
Therefore, we use a learnable fill token $\mathbf{t}$ to replace the masked content. 
The masked image can be written as

\begin{equation}
\tilde{\mathbf{x}} = \mathbf{x}\odot \mathbf{V} + \mathbf{t}\odot \mathbf{M}^{\uparrow},
\end{equation}
where $\odot$ denotes multiplication at corresponding spatial positions. 
In this formulation, the visible part of the original image is preserved, and the masked part is filled with a trainable placeholder. 
This gives the network a more complete image of the inputs. 
The resulting image $\tilde{\mathbf{x}}$ is then passed to the reconstruction pipeline.

\subsubsection{Sparse Convolutional Encoding with Visibility Guidance}
% 掩码感知的下采样与稀疏卷积编码

To adapt masked generative pretraining to a CNN backbone, we introduce a sparse encoding strategy with mask guidance in Stage-A. 
For PCB images, defect recognition often depends on structural irregularities and topology changes, rather than semantic context. 
After masking, the input image is only partly visible. 
If standard dense convolution is used directly, responses from masked regions may be mixed into deeper feature maps. 
This interference can weaken the learning of structural representations from the remaining conductive patterns.

\begin{figure}[t]
  \centering
  \includegraphics[width=1.0\textwidth]{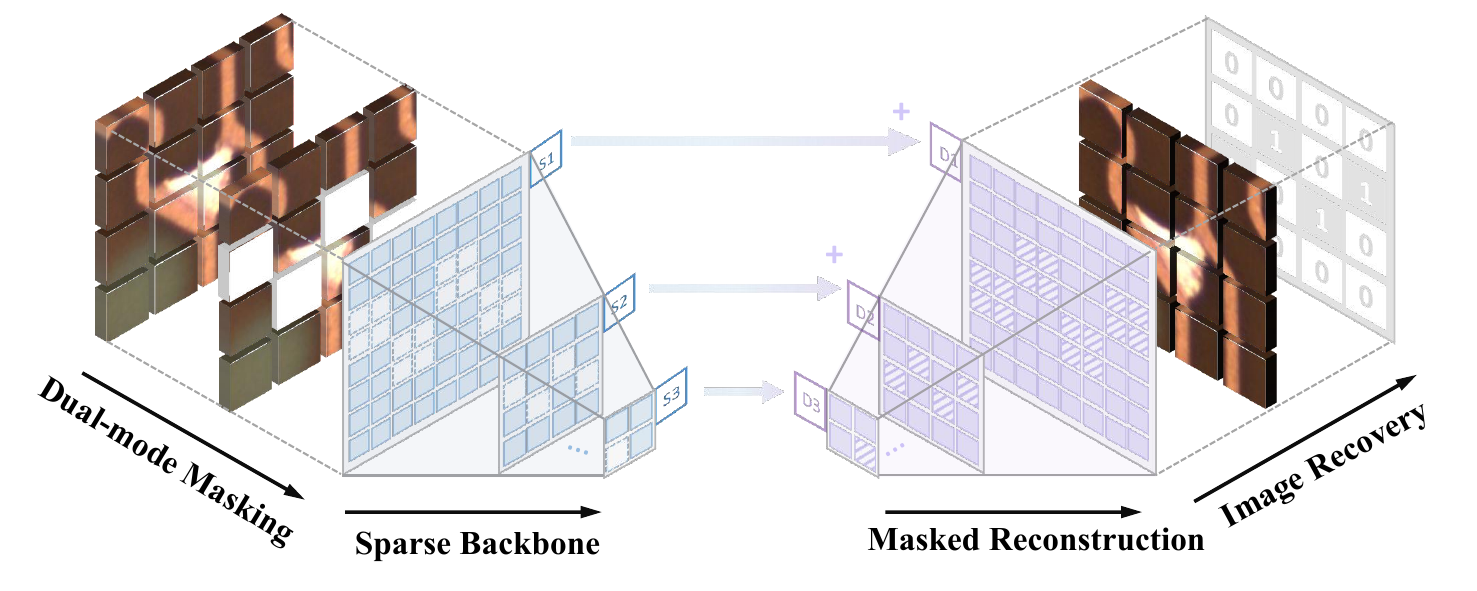}
  \caption{Illustration of the sparse convolutional masked reconstruction process in the pretrain stage. The visible mask is progressively propagated through the YOLOv8 feature hierarchy to suppress invalid responses from masked regions at different feature levels. The resulting sparse hierarchical features are then aggregated by the reconstruction branch to recover missing PCB structures from the remaining visible context.}
  \label{fig:mask_downsample}
\end{figure}

As illustrated in Fig.~\ref{fig:mask_downsample}, the masked image $\tilde{\mathbf{x}}$ and its visible mask $\mathbf{V}$ are processed together by the YOLOv8 encoder. 
We do not explicitly extract only the visible patches. 
Instead, we keep the original convolutional hierarchy and propagate the visible mask along the feature pyramid. 
At each stage, the mask is resized to match the current feature resolution. 
It is used to suppress the responses at masked spatial locations. 
In this manner, invalid features are gradually reduced during hierarchical encoding.

Formally, let $\mathbf{F}^{(l)}$ denote the feature map at stage $l$, and let $\mathbf{V}^{(l)}$ denote the visible mask resized to the same spatial size. 
The mask guided update is defined as
\begin{equation}
\begin{aligned}
\mathbf{Z}^{(l+1)} &= \mathcal{C}^{(l)}\!\left(\mathbf{F}^{(l)} \odot \mathbf{V}^{(l)}\right), \\
\mathbf{V}^{(l+1)} &= \mathcal{R}\!\left(\mathbf{V}^{(l)}\right), \\
\mathbf{F}^{(l+1)} &= \mathbf{Z}^{(l+1)} \odot \mathbf{V}^{(l+1)},
\end{aligned}
\label{eq:mask_sparse_update}
\end{equation}
where $\mathcal{C}^{(l)}(\cdot)$ denotes the transformation of the $l$-th encoder block, $\mathcal{R}(\cdot)$ denotes mask resizing along the hierarchy, and $\odot$ denotes multiplication at corresponding spatial positions.

Based on this sparse encoder, three feature maps with mask guidance are extracted from different stages of the backbone. 
These features retain information from visible regions and suppress responses from masked locations. 
This provides cleaner inputs for the following reconstruction head. 
As a result, the decoder mainly uses features extracted from visible PCB regions. Masked locations have already been suppressed in the encoder, so their corrupted responses are less likely to interfere with the reconstruction of missing conductive structures.

\subsubsection{Multi-scale Reconstruction with Optimization on Masked Regions}

As illustrated in Fig.~\ref{fig:StageA}(b), the masked PCB image is first passed to a sparse YOLOv8s backbone encoder. 
To keep the same structure as the downstream detector, we extract three feature maps from the backbone, denoted as $\{\mathbf{P}_3,\mathbf{P}_4,\mathbf{P}_5\}$. 
These feature maps also provide structural information at different scales. 
For an input of size $640\times640$, they are defined as
\begin{equation}
\mathbf{P}_3 \in \mathbb{R}^{128\times80\times80}, \quad
\mathbf{P}_4 \in \mathbb{R}^{256\times40\times40}, \quad
\mathbf{P}_5 \in \mathbb{R}^{512\times20\times20}.
\end{equation}
Here, $\mathbf{P}_3$ captures fine local details, $\mathbf{P}_4$ describes structural patterns at a middle scale, and $\mathbf{P}_5$ provides broader contextual information.

To fuse these features, we first use lateral $1\times1$ convolutions to project them to the same hidden dimension $C$:
\begin{equation}
\mathbf{P}_3' \in \mathbb{R}^{C\times80\times80}, \quad
\mathbf{P}_4' \in \mathbb{R}^{C\times40\times40}, \quad
\mathbf{P}_5' \in \mathbb{R}^{C\times20\times20},
\end{equation}
where $C=256$ in our implementation. 
After channel alignment, the reconstruction head fuses these features in a lightweight manner. 
First, $\mathbf{P}_5'$ is upsampled and concatenated with $\mathbf{P}_4'$. 
Subsequently, the concatenated feature is processed by a fusion block and residual refinement to obtain the middle scale feature:
\begin{equation}
\mathbf{F}_{mid} = \mathcal{R}_{mid}\Big(\mathcal{F}_{mid}\big([\mathbf{P}_4',\ \mathrm{Up}(\mathbf{P}_5')]\big)\Big),
\end{equation}

Here, $[\mathbf{P}_4',\ \mathrm{Up}(\mathbf{P}_5')]$ means that the aligned feature $\mathbf{P}_4'$ is concatenated with the upsampled feature $\mathbf{P}_5'$ along the channel dimension. 
The upsampling operation $\mathrm{Up}(\mathbf{P}_5')$ uses bilinear interpolation to match the spatial size of $\mathbf{P}_4'$. 
The concatenated feature is then passed through the fusion block $\mathcal{F}_{mid}$ and the residual refinement module $\mathcal{R}_{mid}$ to obtain $\mathbf{F}_{mid}$.

Then, $\mathbf{F}_{mid}$ is upsampled and concatenated with $\mathbf{P}_3'$ to produce the low-scale feature:
\begin{equation}
\mathbf{F}_{low} = \mathcal{R}_{low}\Big(\mathcal{F}_{low}\big([\mathbf{P}_3',\ \mathrm{Up}(\mathbf{F}_{mid})]\big)\Big).
\end{equation}

\begin{figure}[t]
  \centering
  \includegraphics[width=1.0\textwidth]{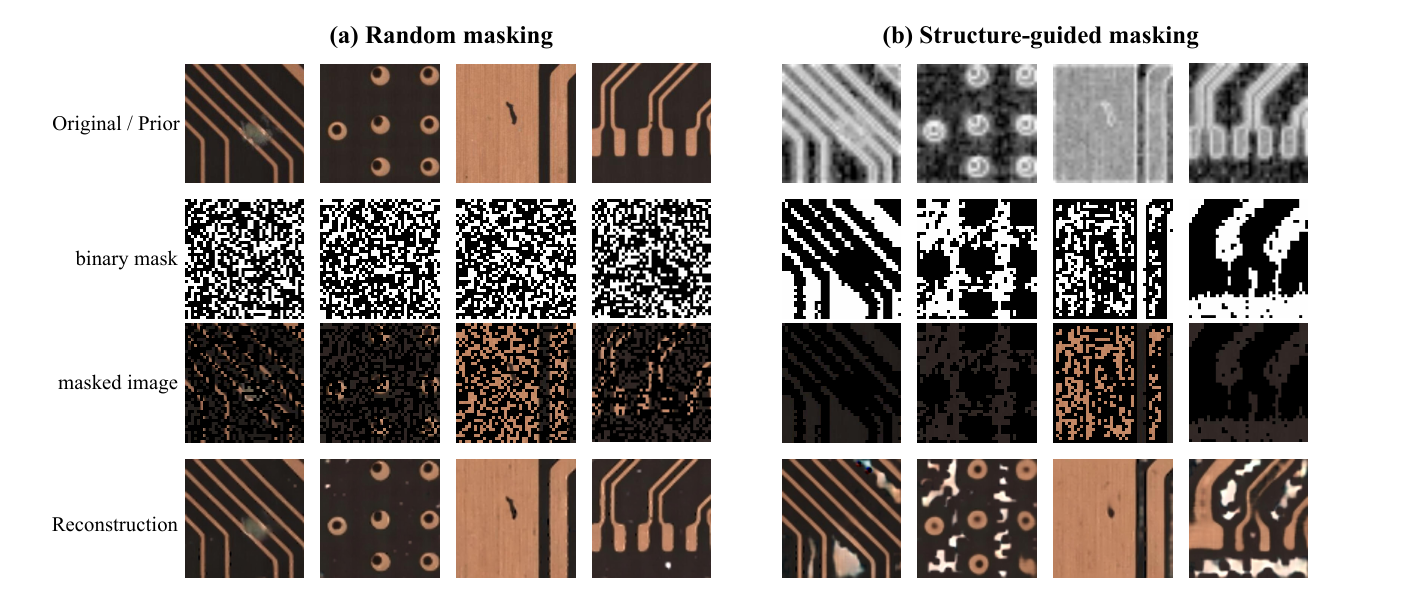}
  \caption{Qualitative comparison between random masking and structure-guided masking in the pretraining stage. Structure-guided masking emphasizes structurally important PCB regions but may make reconstruction overly difficult due to its concentration on key structures. Random masking provides complementary spatial diversity, so the mixed masking strategy balances structural emphasis and reconstruction stability.}
  \label{fig:StageA_vis}
\end{figure}

After this step, The fused feature $\mathbf{F}_{low}$ is processed by a shared trunk and an image prediction branch. 
This produces a coarse reconstruction with three channels. 
The coarse result is finally resized to the input resolution by bilinear interpolation. 
With this design, the decoder remains lightweight while still combining broad contextual cues with fine local structures.

Fig.~\ref{fig:StageA_vis} visualizes the application of random masking and structure-guided masking in Stage-A pretraining, together with the corresponding masked inputs and reconstruction results. 
It can be seen that random masking usually produces smoother reconstruction results. This is mainly because random masks are spatially scattered, making the recovery task relatively easier. 
In contrast, structure-guided masking places more masks on structurally important regions, such as conductive traces, boundaries, and local geometric patterns. This makes the reconstruction task harder, so the recovered image may appear less complete in some local areas. 
However, this does not mean that structure-guided masking is ineffective. 
By forcing the model to recover more informative PCB structures, it encourages the backbone to focus on regions that are more important for downstream defect detection. But relying only on structure-guided masking may make the pretraining task too biased toward high response structures and reduce spatial diversity. Random masking lacks structural guidance, yet it still provides useful diversity for pretraining. For this reason, the mixed masking strategy is used to balance structural emphasis and masking diversity.

After obtaining the reconstructed image $\hat{\mathbf{x}}$, we optimize the network with a reconstruction objective defined on the masked regions. 
This objective is designed to compare the reconstructed pixels with the original image only at masked locations. The reconstruction loss is computed as
\begin{equation}
\mathcal{L}_{rec}
=
\frac{\left\| \mathbf{M}^{\uparrow}\odot(\hat{\mathbf{x}}-\mathbf{x}) \right\|_{1}}
{\sum \mathbf{M}^{\uparrow} + \epsilon},
\end{equation}
where $\epsilon$ is a small constant used to avoid division by zero. 
An $\mathcal{L}_2$ variant can also be used in practice. 
The main principle is the same: only masked pixels contribute to the reconstruction loss, while visible pixels are not included.

Overall, Stage-A learns PCB specific structural priors through structure-guided masking, sparse encoding, and reconstruction on masked regions. 
It also helps the backbone capture contextual relations among local PCB patterns from unlabeled images. After pretraining, the reconstruction head is discarded. Only the pretrained backbone is retained for downstream detection.

\subsection{Stage-B: Transfer to Detection and Supervised Fine-Tuning}
% \subsection{Stage-B：迁移至检测与监督微调}

\begin{figure}[!htb]
  \centering
  \includegraphics[width=1.0\textwidth]{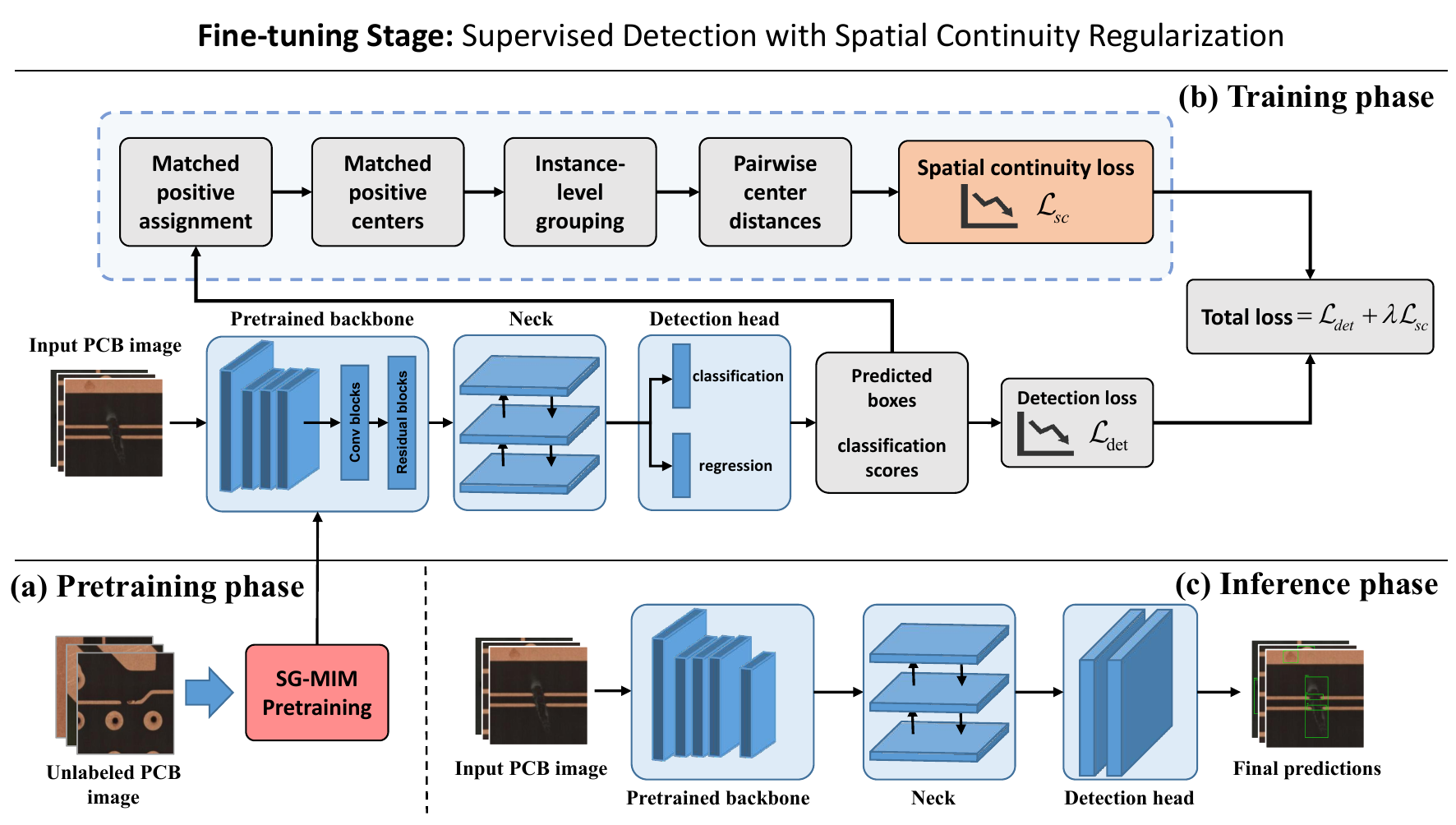}
  \caption{Overview of the fine-tuning stage with spatial continuity regularization. (a) The SG-MIM pretrained backbone is transferred to YOLOv8s for downstream defect detection. (b) During supervised fine-tuning, the spatial continuity loss constrains matched positive predictions assigned to the same defect instance. (c) During inference, the standard detection pipeline is retained without introducing extra computational cost.}
  \label{fig:stageB_framework}
\end{figure}

\subsubsection{Transfer of the Pretrained Backbone}
As illustrated in Fig.~\ref{fig:stageB_framework}(a) and (b), the reconstruction components are removed after Stage-A pretraining. 
Only the pretrained YOLOv8 backbone is retained for downstream detection. 
The encoder used in SG-MIM has the same structure as the native backbone of the detector. Thus, its learned parameters can be directly transferred without an extra adaptation module. The pretrained backbone is combined with the standard YOLOv8 neck and decoupled detection head. 
This forms the Stage-B detector for supervised fine-tuning. In this setting, the structural knowledge learned in Stage-A provides a better initialization for PCB defect detection.

\subsubsection{Spatial Continuity Regularization}

In addition to the standard detection branch, we employ a spatial continuity regularization term during Stage-B training, as shown in Fig.~\ref{fig:stageB_framework}(b). 
The motivation is straightforward. 
Although the pretrained backbone improves feature representation, the detector may still produce fragmented or scattered boxes for elongated PCB defects. 
This problem is more evident for defect categories with slender, connected, or locally continuous shapes.

To reduce this problem, the regularization term constrains the spatial distribution of matched positive predictions. 
Here, matched positive predictions refer to the predictions selected as positive samples after label assignment.
As shown in Fig.~\ref{fig:stageB_framework}(b), the detector first performs positive sample assignment. 
It then extracts the centers of the matched positive predictions. 
These centers are grouped according to their assigned ground-truth instances. 
Finally, pairwise center distances are computed within each instance-level group. 
This process measures how scattered the positive predictions assigned to the same defect instance are. 
During optimization, overly dispersed positive predictions within the same instance-level group are identified. 
These dispersed predictions are then penalized.

For the $k$-th ground-truth instance in image $b$, let $\mathcal{P}_{b,k}=\{\mathbf{p}_i\}_{i=1}^{N_{b,k}}$ denote the set of center points of all matched positive predictions assigned to this instance, where
\begin{equation}
\mathbf{p}_i = (x_i, y_i).
\end{equation}
The center coordinates are normalized by the image size to reduce the effect of image scale. 
The spatial continuity regularization is defined as
\begin{equation}
\mathcal{L}_{sc}
=
\frac{1}{B}
\sum_{b=1}^{B}
\frac{1}{K_b}
\sum_{k=1}^{K_b}
\frac{1}{|\Omega_{b,k}|}
\sum_{(i,j)\in\Omega_{b,k}}
\left\|
\mathbf{p}_i - \mathbf{p}_j
\right\|_2,
\end{equation}
where $B$ is the batch size, $K_b$ is the number of ground-truth instances in image $b$, and $\Omega_{b,k}$ denotes the set of all unordered center pairs in $\mathcal{P}_{b,k}$. 
If fewer than two matched positive predictions are available for a ground-truth instance, the corresponding term is set to zero.

This loss does not replace the conventional box regression objective. 
It is used as an auxiliary constraint on the spatial arrangement of positive predictions assigned to the same defect instance. 
A smaller $\mathcal{L}_{sc}$ means that positive predictions assigned to the same instance tend to form a more spatially coherent pattern.
A larger $\mathcal{L}_{sc}$ indicates a more scattered prediction pattern. 
By adding this constraint, the detector is encouraged to produce more coherent localization results and fewer fragmented responses on continuous defect regions.

\subsubsection{Overall Loss, Training, and Inference}

The final optimization objective of Stage-B is given by
\begin{equation}
\mathcal{L} = \mathcal{L}_{det} + \lambda\,\mathcal{L}_{sc},
\end{equation}
where $\mathcal{L}_{det}$ denotes the standard YOLOv8 detection loss and $\lambda$ controls the contribution of the spatial continuity regularization. 
As summarized in Fig.~\ref{fig:stageB_framework}(b), the spatial continuity branch is used only during supervised training. 
It is optimized together with the native detection objective.

During inference, only the standard YOLOv8 detection pipeline is kept, as shown in Fig.~\ref{fig:stageB_framework}(c). 
The pretrained backbone, neck, and detection head are directly used to generate final predictions. 
This regularization term is used only during training and is not involved in inference. 
Together with the Stage-A pretraining process in Fig.~\ref{fig:stageB_framework}(a), this forms a complete two stage framework. 
It improves the spatial coherence of detection results without adding extra computation during inference.

\section{Experiments and Results}
\subsection{Experimental Settings}
\subsubsection{Experimental Environment and Training Strategy}

All experiments are conducted on a workstation running Ubuntu 20.04.5 LTS, equipped with an Intel(R) Xeon(R) Platinum 8474C CPU, 1.0~TB RAM, and an NVIDIA GeForce RTX 4090 D GPU with 24~GB memory. The software environment includes NVIDIA driver 580.76.05, CUDA 13.0 (driver-reported), Python 3.8.10, PyTorch 2.0.0+cu118, PyTorch CUDA 11.8, and Ultralytics 8.4.7.
% 所有实验均在一台运行 Ubuntu 20.04.5 LTS 的工作站上进行，该设备配备 Intel(R) Xeon(R) Platinum 8474C 处理器、1.0~TB 内存以及一块 24~GB 显存的 NVIDIA GeForce RTX 4090 D GPU。
% 软件环境包括 NVIDIA driver 580.76.05、CUDA 13.0（驱动报告版本）、Python 3.8.10、PyTorch 2.0.0+cu118、PyTorch CUDA 11.8 以及 Ultralytics 8.4.7。

The proposed method is trained in two stages, including Stage-A self-supervised pretraining and Stage-B supervised detection fine-tuning. In Stage-A, structure-guided masked image modeling is performed on unlabeled PCB images with an input size of $640 \times 640$, a patch size of 16, and a masking ratio of 0.60. The model is trained for 400 epochs using AdamW with a batch size of 64, an initial learning rate of $8\times10^{-4}$, and a weight decay of $10^{-4}$.
% 所提出的方法分两个阶段进行训练，包括 Stage-A 自监督预训练和 Stage-B 监督检测微调。
% 在 Stage-A 中，我们在无标注 PCB 图像上执行结构引导的掩码图像建模，输入尺寸为 $640 \times 640$，patch 大小为 16，掩码比例为 0.60。
% 模型采用 AdamW 进行优化，训练 400 个 epoch，batch size 为 64，初始学习率为 $8\times10^{-4}$，权重衰减为 $10^{-4}$。

In Stage-B, the pretrained backbone is transferred to YOLOv8 for downstream defect detection. The detector is trained for 100 epochs with an input size of 640 and a batch size of 16. AdamW is used as the optimizer with an initial learning rate of $5\times10^{-4}$ and a weight decay of $5\times10^{-4}$. Unless otherwise specified, the spatial continuity regularization coefficient $\lambda$ is set to 0.20.
% 在 Stage-B 中，预训练得到的骨干网络被迁移到 YOLOv8 中用于下游缺陷检测。
% 检测器训练 100 个 epoch，输入尺寸为 640，batch size 为 16。
% 采用 AdamW 作为优化器，初始学习率为 $5\times10^{-4}$，权重衰减为 $5\times10^{-4}$。
% 除非特别说明，空间连续性正则化系数 $\lambda$ 设为 0.20。

\subsubsection{Dataset}

\begin{figure}[!htbp]
    \centering
    \includegraphics[width=\columnwidth]{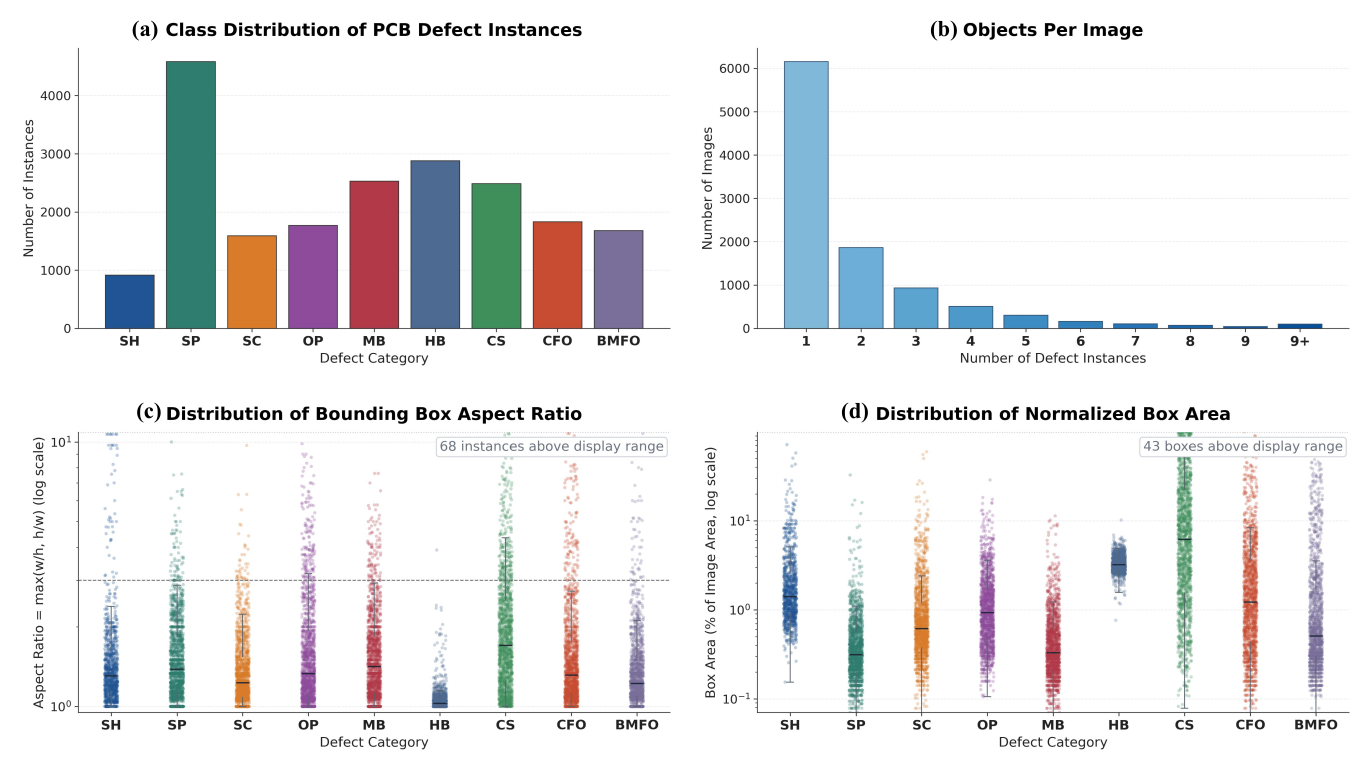}
    \caption{Statistical analysis of the DsPCBSD+ dataset, indicating imbalanced defect categories, predominantly single defect images, and diverse defect shapes and scales.}
    \label{fig:dataset_distribution}
\end{figure}

The experiments are conducted on the DsPCBSD+ dataset~\cite{lv2024dataset}, a benchmark dataset for PCB surface defect detection. It contains nine defect categories, including Short (SH), Spur (SP), Spurious copper (SC), Open (OP), Mouse bite (MB), Hole breakout (HB), Conductor scratch (CS), Conductor foreign object (CFO), and Base material foreign object (BMFO). Following the official setting of the dataset, all labeled images are randomly divided into training and validation subsets at a ratio of 8:2. The training subset contains 8,208 images with 16,184 defect annotations, while the validation subset contains 2,051 images with 4,092 annotations. All reported results in this work are evaluated on the validation subset.

Fig.~\ref{fig:dataset_distribution} summarizes the statistical characteristics of the dataset. As shown in Fig.~\ref{fig:dataset_distribution}(a), the number of defect instances varies noticeably across categories. SP has the largest number of instances, whereas SH contains relatively fewer samples, indicating a certain degree of class imbalance. Fig.~\ref{fig:dataset_distribution}(b) further shows the distribution of the number of defect instances per image. Most images contain only one defect instance, while a smaller portion contains multiple defects.This distribution indicates that most PCB images contain only a limited number of defect instances, while the detector still needs to handle cases where several defects appear simultaneously.

The geometric properties of defect instances are shown in Fig.~\ref{fig:dataset_distribution}(c) and Fig.~\ref{fig:dataset_distribution}(d). The aspect-ratio distribution in Fig.~\ref{fig:dataset_distribution}(c) differs clearly among categories. In particular, CS and CFO contain many instances with large aspect ratios, indicating that these defects often appear as slender or elongated regions. Such shapes are closely related to the fragmented prediction problem discussed earlier, where a continuous defect may be split into several scattered bounding boxes. Meanwhile, Fig.~\ref{fig:dataset_distribution}(d) shows that the normalized box areas span a wide range. SP and MB are mainly concentrated in small-area regions, suggesting that many of their instances occupy only a limited portion of the image and are therefore more vulnerable to weak visual responses and background interference. In contrast, HB and part of CS cover relatively larger areas, showing that the dataset contains defects at different spatial scales.

These dataset characteristics are consistent with the practical challenges addressed in this work. PCB defects often appear on highly regular traces, pads, and junction regions, and their discriminative cues are more related to structural continuity and local topology than to high-level semantic appearance. Therefore, small defects require stronger structural representation under weak contrast, while elongated defects require more coherent localization. These properties make DsPCBSD+ suitable for evaluating both the representation learning ability and the localization robustness of the proposed method.

\subsubsection{Evaluation Metrics}

To quantitatively evaluate the detection performance, Precision ($P$), Recall ($R$), $\mathrm{mAP}_{0.5}$, and $\mathrm{mAP}_{0.5:0.95}$ are used as the main evaluation metrics.

Precision and Recall are defined as
\begin{equation}
P=\frac{TP}{TP+FP}
\end{equation}
\begin{equation}
R=\frac{TP}{TP+FN}
\end{equation}
where $TP$, $FP$, and $FN$ denote the numbers of true positives, false positives, and false negatives, respectively.

For each defect category, the average precision ($AP$) is computed from the area under the precision--recall curve:
\begin{equation}
AP=\int_{0}^{1} P(R)\, dR
\end{equation}
where $P(R)$ denotes precision as a function of recall.

The mean average precision ($mAP$) is obtained by averaging the $AP$ values over all defect categories:
\begin{equation}
mAP=\frac{1}{N}\sum_{i=1}^{N} AP_i
\end{equation}
where $N$ is the number of defect categories. In this work, $\mathrm{mAP}_{0.5}$ denotes the mean average precision at an IoU threshold of 0.5, while $\mathrm{mAP}_{0.5:0.95}$ denotes the average mAP over IoU thresholds from 0.5 to 0.95 with a step size of 0.05.

\subsection{Results and Analysis}
\subsubsection{Training Convergence Analysis}
\begin{figure}[t]
    \centering
    \includegraphics[width=\textwidth]{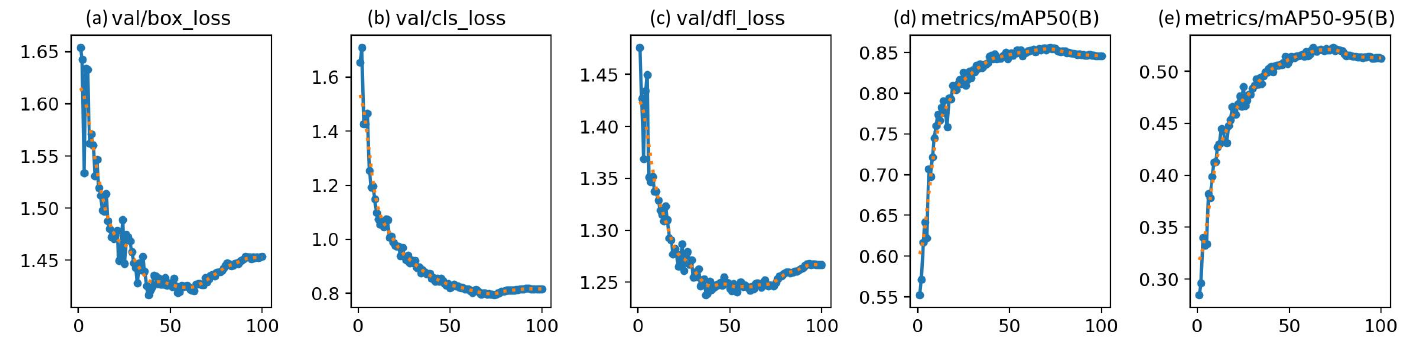}
   \caption{Validation curves of the proposed method on the DsPCBSD+ dataset.}
    \label{fig:training_curves}
\end{figure}

Fig.~\ref{fig:training_curves} shows the validation curves of the proposed method. 
From Fig.~\ref{fig:training_curves}(a), (b), and (c), it can be observed that the validation box loss, classification loss, and DFL loss all drop rapidly in the early stage, especially within the first 20 epochs. After about 30--50 epochs, the three loss curves become much more stable and remain in a relatively narrow range. 
Although slight fluctuations can still be observed in the later stage, no obvious divergence appears throughout training. 
Fig.~\ref{fig:training_curves}(d) and (e) show that both $\mathrm{mAP}_{0.5}$ and $\mathrm{mAP}_{0.5:0.95}$ increase quickly in the early stage and gradually approach saturation after about 50--70 epochs.

These curves indicate that the model enters an effective optimization regime at a relatively early stage. 
This suggests that the pretrained initialization helps the detector converge more efficiently during fine-tuning. 
In addition, the continued improvement of $\mathrm{mAP}_{0.5:0.95}$ before saturation shows that the method improves not only detection performance but also localization quality under stricter IoU thresholds. 

\subsubsection{Detection Performance for Different Defect Categories}
\begin{table}[t]
\centering
\caption{Detection performance before and after introducing the proposed improvements for each defect category on the DsPCBSD+ validation set.}
\label{tab:per_class_results_compare}
\resizebox{1.0\textwidth}{!}{
\begin{tabular}{lcccccccc}
\hline
\multirow{2}{*}{Defect} 
& \multicolumn{2}{c}{Precision (\%)} 
& \multicolumn{2}{c}{Recall (\%)} 
& \multicolumn{2}{c}{$\mathrm{AP}_{0.5}$ (\%)} 
& \multicolumn{2}{c}{$\mathrm{AP}_{0.5:0.95}$ (\%)} \\
\cline{2-9}
& Before & After & Before & After & Before & After & Before & After \\
\hline
SH   & 84.3 & 83.1{\scriptsize\textcolor{blue}{$\,\downarrow$1.2}} & 88.7 & 85.8{\scriptsize\textcolor{blue}{$\,\downarrow$2.9}} & 89.3 & 91.0{\scriptsize\textcolor{red}{$\,\uparrow$1.7}} & 58.5 & 60.3{\scriptsize\textcolor{red}{$\,\uparrow$1.8}} \\
SP   & 84.0 & 84.2{\scriptsize\textcolor{red}{$\,\uparrow$0.2}} & 76.7 & 77.8{\scriptsize\textcolor{red}{$\,\uparrow$1.1}} & 84.5 & 85.5{\scriptsize\textcolor{red}{$\,\uparrow$1.0}} & 38.8 & 40.0{\scriptsize\textcolor{red}{$\,\uparrow$1.2}} \\
SC   & 79.1 & 79.0{\scriptsize\textcolor{blue}{$\,\downarrow$0.1}} & 79.3 & 81.8{\scriptsize\textcolor{red}{$\,\uparrow$2.5}} & 81.3 & 84.5{\scriptsize\textcolor{red}{$\,\uparrow$3.2}} & 50.6 & 53.2{\scriptsize\textcolor{red}{$\,\uparrow$2.6}} \\
OP   & 78.6 & 84.5{\scriptsize\textcolor{red}{$\,\uparrow$5.9}} & 85.5 & 85.7{\scriptsize\textcolor{red}{$\,\uparrow$0.2}} & 89.1 & 89.2{\scriptsize\textcolor{red}{$\,\uparrow$0.1}} & 54.6 & 54.6{\scriptsize\textcolor{red}{$\,\uparrow$0.0}} \\
MB   & 84.6 & 84.2{\scriptsize\textcolor{blue}{$\,\downarrow$0.4}} & 75.3 & 77.8{\scriptsize\textcolor{red}{$\,\uparrow$2.5}} & 83.4 & 84.9{\scriptsize\textcolor{red}{$\,\uparrow$1.5}} & 39.2 & 41.8{\scriptsize\textcolor{red}{$\,\uparrow$2.6}} \\
HB   & 93.2 & 93.3{\scriptsize\textcolor{red}{$\,\uparrow$0.1}} & 94.5 & 95.4{\scriptsize\textcolor{red}{$\,\uparrow$0.9}} & 98.4 & 98.3{\scriptsize\textcolor{blue}{$\,\downarrow$0.1}} & 84.3 & 83.9{\scriptsize\textcolor{blue}{$\,\downarrow$0.4}} \\
CS   & 73.1 & 70.7{\scriptsize\textcolor{blue}{$\,\downarrow$2.4}} & 63.6 & 68.5{\scriptsize\textcolor{red}{$\,\uparrow$4.9}} & 70.9 & 73.7{\scriptsize\textcolor{red}{$\,\uparrow$2.8}} & 42.6 & 45.1{\scriptsize\textcolor{red}{$\,\uparrow$2.5}} \\
CFO  & 74.3 & 71.9{\scriptsize\textcolor{blue}{$\,\downarrow$2.4}} & 64.8 & 68.3{\scriptsize\textcolor{red}{$\,\uparrow$3.5}} & 71.8 & 73.3{\scriptsize\textcolor{red}{$\,\uparrow$1.5}} & 39.7 & 43.4{\scriptsize\textcolor{red}{$\,\uparrow$3.7}} \\
BMFO & 86.3 & 83.3{\scriptsize\textcolor{blue}{$\,\downarrow$3.0}} & 85.0 & 86.1{\scriptsize\textcolor{red}{$\,\uparrow$1.1}} & 88.0 & 89.3{\scriptsize\textcolor{red}{$\,\uparrow$1.3}} & 47.5 & 48.3{\scriptsize\textcolor{red}{$\,\uparrow$0.8}} \\
\hline
\textbf{All classes} 
& \textbf{81.9} 
& \textbf{81.6}{\scriptsize\textcolor{blue}{\textbf{$\,\downarrow$0.3}}} 
& \textbf{79.3} 
& \textbf{80.8}{\scriptsize\textcolor{red}{\textbf{$\,\uparrow$1.5}}} 
& \textbf{84.1} 
& \textbf{85.5}{\scriptsize\textcolor{red}{\textbf{$\,\uparrow$1.4}}} 
& \textbf{50.7} 
& \textbf{52.3}{\scriptsize\textcolor{red}{\textbf{$\,\uparrow$1.6}}} \\
\hline
\end{tabular}
}
\end{table}
To provide a more detailed class-level evaluation, we further compare the detection performance for each defect category before and after introducing the proposed improvements. The corresponding results are summarized in Table~\ref{tab:per_class_results_compare}.
% 为了提供更细粒度的类别级评估，我们进一步比较了引入所提出改进前后各缺陷类别的检测性能。
% 对应结果汇总于表~\ref{tab:per_class_results_compare} 中。

For SC, the improved model shows a better ability to capture structurally complex defect patterns, with recall increasing from 79.3\% to 81.8\%, $\mathrm{AP}_{0.5}$ rising from 81.3\% to 84.5\%, and $\mathrm{AP}_{0.5:0.95}$ improving from 50.6\% to 53.2\%. For CS and CFO, the gains under the stricter $\mathrm{AP}_{0.5:0.95}$ metric are more evident. CS improves from 42.6\% to 45.1\%, and CFO improves from 39.7\% to 43.4\%. These changes indicate more stable localization on difficult categories with stronger structural interference.

Another interesting observation can be made from Table~\ref{tab:per_class_results_compare}. 
For most defect categories, Precision decreases slightly, while Recall increases more consistently. 
This change still leads to overall gains in $\mathrm{AP}_{0.5}$ and $\mathrm{AP}_{0.5:0.95}$. 
It suggests that the baseline detector is relatively conservative and tends to miss some true defect instances. 
After the proposed improvements, the detector becomes more willing to respond to potential defect regions. More true positives are recovered, even though a small amount of extra false positives may also be introduced. 

In general, $\mathrm{mAP}_{0.5}$ rises from 84.1\% to 85.5\%, and $\mathrm{mAP}_{0.5:0.95}$ improves from 50.7\% to 52.3\%. 
The proposed improvements make the detector more robust across categories, even though the gains are not equally large for every defect type.

\begin{table}[t]
\centering
\caption{Performance comparison with benchmark models on the DsPCBSD+ validation set.}
\label{tab:main_results}
\begin{tabular}{lcc}
\hline
Method & $\mathrm{mAP}_{0.5}$ (\%) & $\mathrm{mAP}_{0.5:0.95}$ (\%) \\
\hline
PRB-FPN-CSP~\cite{chen2021parallel} & 81.0{\scriptsize\textcolor{blue}{$\pm$0.4}} & 45.3{\scriptsize\textcolor{blue}{$\pm$0.1}} \\
PPYOLOEs~\cite{xu2022pp} & 82.7{\scriptsize\textcolor{blue}{$\pm$0.7}} & 46.0{\scriptsize\textcolor{blue}{$\pm$0.3}} \\
YOLOv5s(8.0)~\cite{yolov5} & 84.3{\scriptsize\textcolor{blue}{$\pm$0.5}} & 48.3{\scriptsize\textcolor{blue}{$\pm$0.3}} \\
DAMOYOLOs~\cite{xu2022damo} & 84.8{\scriptsize\textcolor{blue}{$\pm$0.3}} & 48.5{\scriptsize\textcolor{blue}{$\pm$0.2}} \\
RTMDETs~\cite{lyu2022rtmdet} & 84.8{\scriptsize\textcolor{blue}{$\pm$0.2}} & 48.6{\scriptsize\textcolor{blue}{$\pm$0.3}} \\
RT-DETR~\cite{zhao2024rtdetr} & 84.8{\scriptsize\textcolor{blue}{$\pm$0.3}} & 49.2{\scriptsize\textcolor{blue}{$\pm$0.3}} \\
Co-DETR~\cite{zong2023codetr} & 82.1{\scriptsize\textcolor{blue}{$\pm$0.5}} & 49.6{\scriptsize\textcolor{blue}{$\pm$0.2}} \\
YOLOv6s(3.0)~\cite{li2023efd} & 85.2{\scriptsize\textcolor{blue}{$\pm$0.2}} & 49.7{\scriptsize\textcolor{blue}{$\pm$0.2}} \\
YOLOv8s~\cite{ultralytics2023yolov8docs,ultralytics2023yolo} & 84.1{\scriptsize\textcolor{blue}{$\pm$0.8}} & 50.7{\scriptsize\textcolor{blue}{$\pm$0.6}} \\
YOLOv10~\cite{wang2024yolov10} & 84.7{\scriptsize\textcolor{blue}{$\pm$0.3}} & 51.0{\scriptsize\textcolor{blue}{$\pm$0.2}} \\
YOLOv6-L6~\cite{li2022yolov6} & 85.1{\scriptsize\textcolor{blue}{$\pm$0.1}} & 51.4{\scriptsize\textcolor{blue}{$\pm$0.5}} \\
\hline
\textbf{Ours} &
{\textcolor{red}{\textbf{85.5}}}{\scriptsize\textcolor{blue}{$\pm$0.4}} &
{\textcolor{red}{\textbf{52.3}}}{\scriptsize\textcolor{blue}{$\pm$0.2}} \\
\hline
\end{tabular}
\end{table}

\subsubsection{Comparison with Benchmark Methods}

\begin{figure}[!htb]
    \centering
    \includegraphics[width=1.0\textwidth]{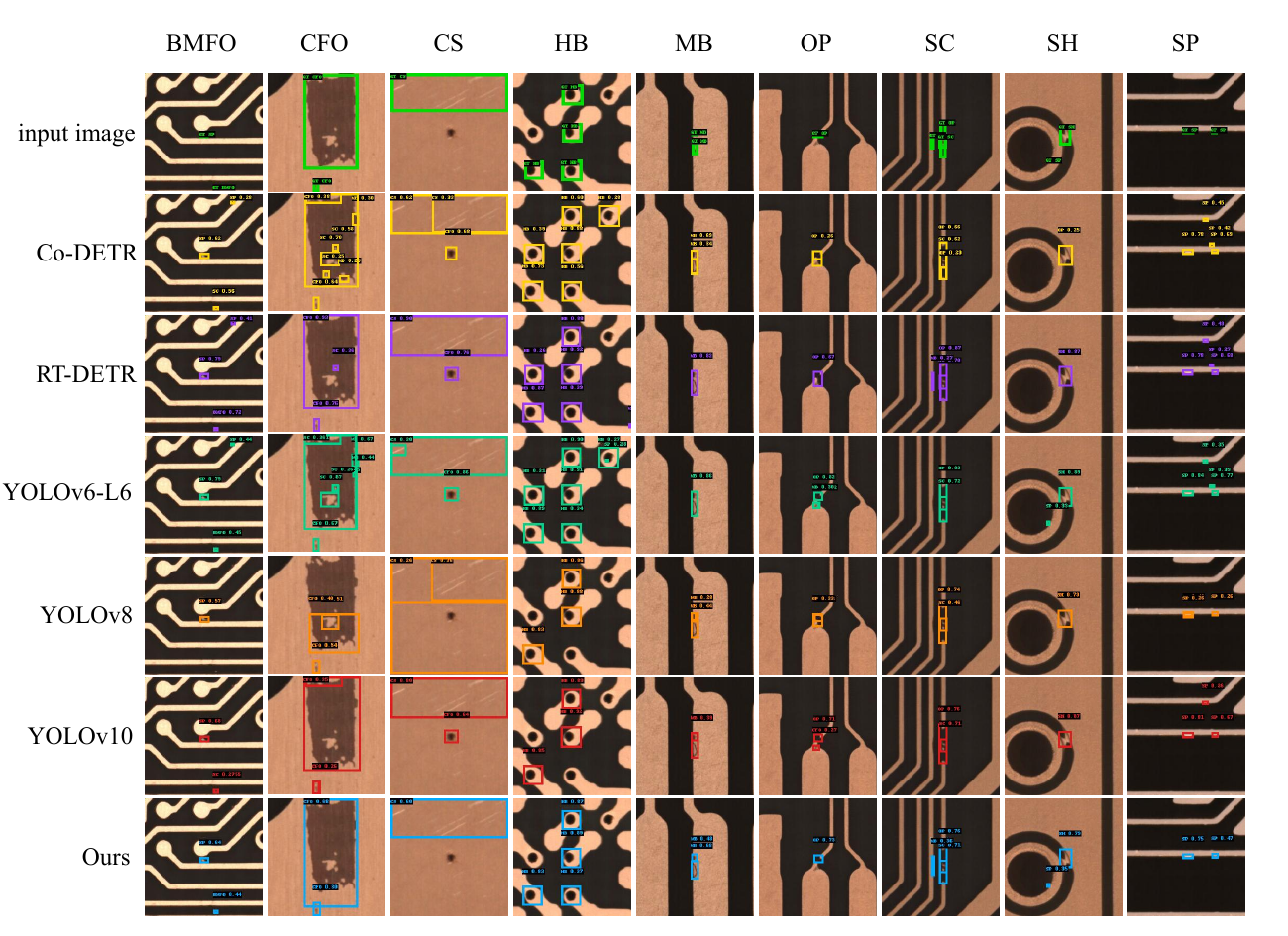}
    \caption{Qualitative comparison of different detectors on representative samples from the DsPCBSD+ dataset.}
    \label{fig:qualitative_results}
\end{figure}

To verify the superiority of our model, we compare it with several detectors under the same experimental setting on the DsPCBSD+ validation set.  The comparison results are shown in Table~\ref{tab:main_results} and Fig.~\ref{fig:qualitative_results}. Due to space limitations, we mainly select detectors with relatively higher $\mathrm{mAP}_{0.5:0.95}$ values for qualitative visualization.
% 为验证所提模型的优越性，我们在相同的实验设置下，于DsPCBSD+验证集上将其与多种先进检测器进行了对比。对比结果如表1和图7所示。由于篇幅所限，定性可视化主要选取了具有较高mAP 
% 0.5:0.95值的检测器。

It can be seen from Table~\ref{tab:main_results} that the proposed method achieves better performance than all compared detectors on the DsPCBSD+ validation set. Our method reaches 85.5\% in $\mathrm{mAP}_{0.5}$ and 52.3\% in $\mathrm{mAP}_{0.5:0.95}$, which are the best results in the table. Compared with YOLOv8s, the proposed method improves $\mathrm{mAP}_{0.5}$ from 84.1\% to 85.5\% and $\mathrm{mAP}_{0.5:0.95}$ from 50.7\% to 52.3\%. It also surpasses strong baselines such as YOLOv10 and YOLOv6-L6, especially on $\mathrm{mAP}_{0.5:0.95}$, which indicates better localization robustness. Compared with RT-DETR and Co-DETR, our method still achieves higher overall accuracy. These results demonstrate the effectiveness of the proposed framework for improving both detection performance and localization quality in PCB defect detection.

As shown in Fig.~\ref{fig:qualitative_results}, the compared detectors exhibit distinct error patterns on challenging samples. 
For CFO, Co-DETR and YOLOv6-L6 tend to produce multiple fragmented responses inside the defective region, while RT-DETR, YOLOv8, and YOLOv10 give relatively coarse predictions. In contrast, our method yields a more compact detection result that better matches the main defect area. For CS and SH, the defect cues are weak and occupy only limited local regions. Several compared detectors either produce incomplete responses or are easily affected by surrounding structures. By comparison, the proposed method gives cleaner and more stable localization, which suggests better sensitivity to subtle defect patterns under complex backgrounds. For SP, several compared detectors still generate multiple neighboring boxes for the same defect pattern, whereas the proposed method maintains more concentrated detections. This again shows an advantage in suppressing redundant predictions on thin and continuous structures. 

\subsubsection{Comparison on Improvement of Typical Challenges}
\begin{figure}[!htb]
    \centering
    \includegraphics[width=0.75\textwidth]{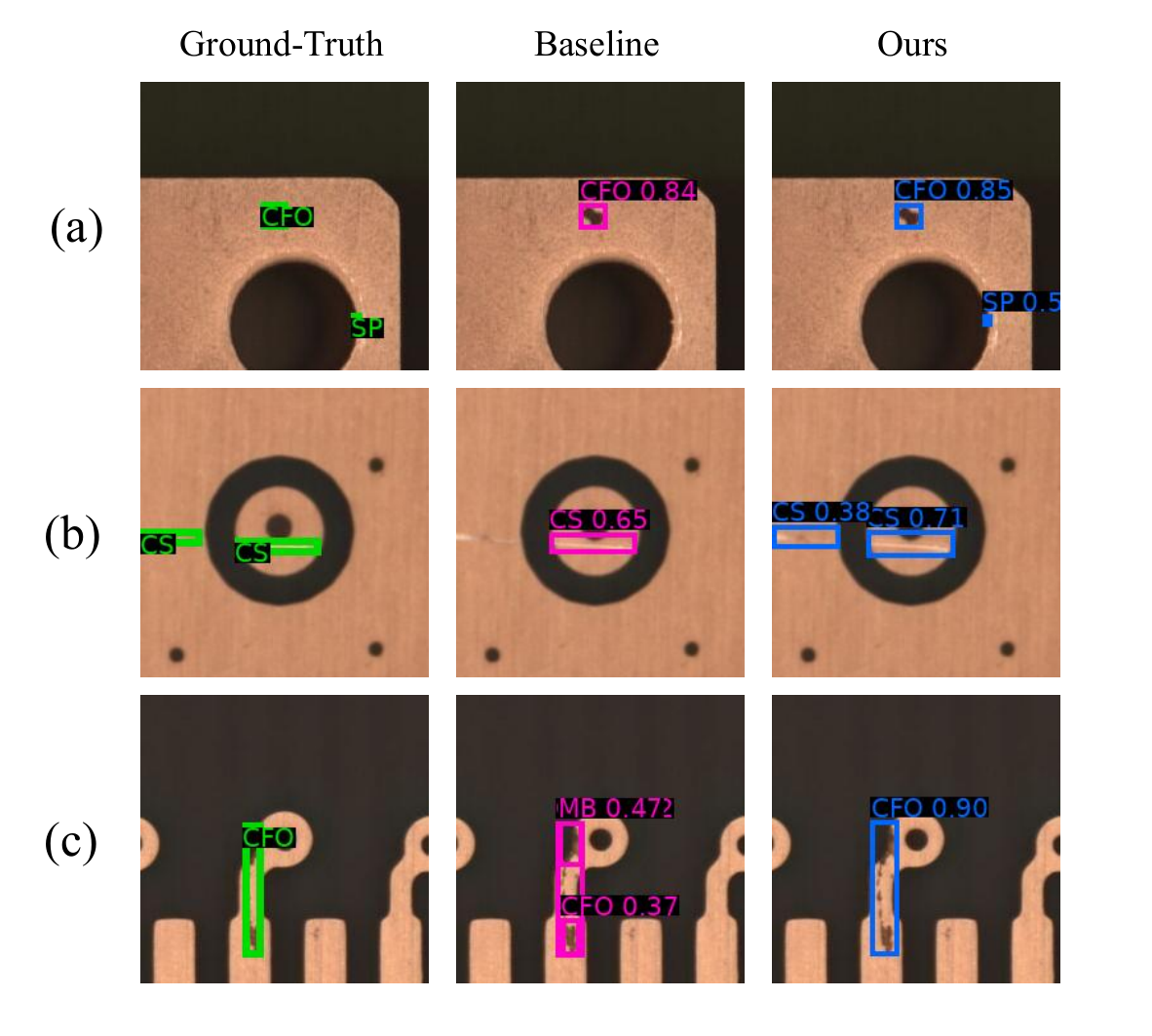}
    \caption{Qualitative comparison on typical challenging samples before and after improvement. }
    \label{fig:qualitative_before_after}
\end{figure}
To more clearly show how our method alleviates the challenges discussed above, Fig.~\ref{fig:qualitative_before_after} presents representative visual comparisons between the baseline detector and the full model on several difficult PCB samples. We select these examples that cover weak tiny defects, locally continuous defect patterns, and elongated defect regions.

As shown in Fig.~\ref{fig:qualitative_before_after}, the baseline tends to make detection errors on challenging defect samples. After the improvements, the detector shows consistently better behavior across all three cases: in Fig.~\ref{fig:qualitative_before_after}(a), the full model successfully recovers the small SP defect that is missed by the baseline while retaining the CFO response; in Fig.~\ref{fig:qualitative_before_after}(b), the baseline fails to capture the full semantic structure of the continuous defect. 
By contrast, our method produces a more complete prediction that better matches the defect extent; and in Fig.~\ref{fig:qualitative_before_after}(c), two separated baseline predictions are effectively consolidated into a more compact and unified localization. These visual results indicate that the proposed designs effectively alleviate the targeted challenges. 
This finding is consistent with the quantitative gains reported in the previous sections.

\begin{table}[t]
\centering
\caption{Ablation results of different components on the DsPCBSD+ validation set.}
\label{tab:component_ablation}
\footnotesize
\renewcommand{\arraystretch}{1.15}
\setlength{\tabcolsep}{3.2pt}

\begin{adjustbox}{width=\textwidth,center}
\begin{tabular}{@{}cccccccc@{}}
\toprule
YOLOv8s &
SG Mixed Mask &
Sparse Conv. &
Spatial Loss &
Precision &
Recall &
$\mathrm{mAP}_{0.5}$ &
$\mathrm{mAP}_{0.5:0.95}$ \\
\midrule
\checkmark &  &  &  & 81.9 & 79.3 & 84.1 & 50.7 \\

\checkmark & \checkmark &  &  &
81.4{\scriptsize\textcolor{blue}{$\,\downarrow$0.5}} &
80.1{\scriptsize\textcolor{red}{$\,\uparrow$0.8}} &
84.6{\scriptsize\textcolor{red}{$\,\uparrow$0.5}} &
51.3{\scriptsize\textcolor{red}{$\,\uparrow$0.6}} \\

\checkmark &  & \checkmark &  &
81.5{\scriptsize\textcolor{blue}{$\,\downarrow$0.4}} &
79.5{\scriptsize\textcolor{red}{$\,\uparrow$0.2}} &
84.2{\scriptsize\textcolor{red}{$\,\uparrow$0.1}} &
50.8{\scriptsize\textcolor{red}{$\,\uparrow$0.1}} \\

\checkmark &  &  & \checkmark &
82.4{\scriptsize\textcolor{red}{$\,\uparrow$0.5}} &
80.4{\scriptsize\textcolor{red}{$\,\uparrow$1.1}} &
85.0{\scriptsize\textcolor{red}{$\,\uparrow$0.9}} &
51.6{\scriptsize\textcolor{red}{$\,\uparrow$0.9}} \\

\checkmark & \checkmark & \checkmark &  &
82.2{\scriptsize\textcolor{red}{$\,\uparrow$0.3}} &
80.6{\scriptsize\textcolor{red}{$\,\uparrow$1.3}} &
85.2{\scriptsize\textcolor{red}{$\,\uparrow$1.1}} &
52.0{\scriptsize\textcolor{red}{$\,\uparrow$1.3}} \\

\checkmark & \checkmark &  & \checkmark &
82.5{\scriptsize\textcolor{red}{$\,\uparrow$0.6}} &
79.2{\scriptsize\textcolor{blue}{$\,\downarrow$0.1}} &
84.8{\scriptsize\textcolor{red}{$\,\uparrow$0.7}} &
51.4{\scriptsize\textcolor{red}{$\,\uparrow$0.7}} \\

\checkmark &  & \checkmark & \checkmark &
81.2{\scriptsize\textcolor{blue}{$\,\downarrow$0.7}} &
80.9{\scriptsize\textcolor{red}{$\,\uparrow$1.6}} &
85.3{\scriptsize\textcolor{red}{$\,\uparrow$1.2}} &
52.2{\scriptsize\textcolor{red}{$\,\uparrow$1.5}} \\

\checkmark & \checkmark & \checkmark & \checkmark &
\textbf{81.6}{\scriptsize\textcolor{blue}{\textbf{$\,\downarrow$0.3}}} &
\textbf{80.8}{\scriptsize\textcolor{red}{\textbf{$\,\uparrow$1.5}}} &
\textbf{85.5}{\scriptsize\textcolor{red}{\textbf{$\,\uparrow$1.4}}} &
\textbf{52.3}{\scriptsize\textcolor{red}{\textbf{$\,\uparrow$1.6}}} \\
\bottomrule
\end{tabular}
\end{adjustbox}
\end{table}

\begin{figure}[t]
    \centering
    \includegraphics[width=1.0\textwidth]{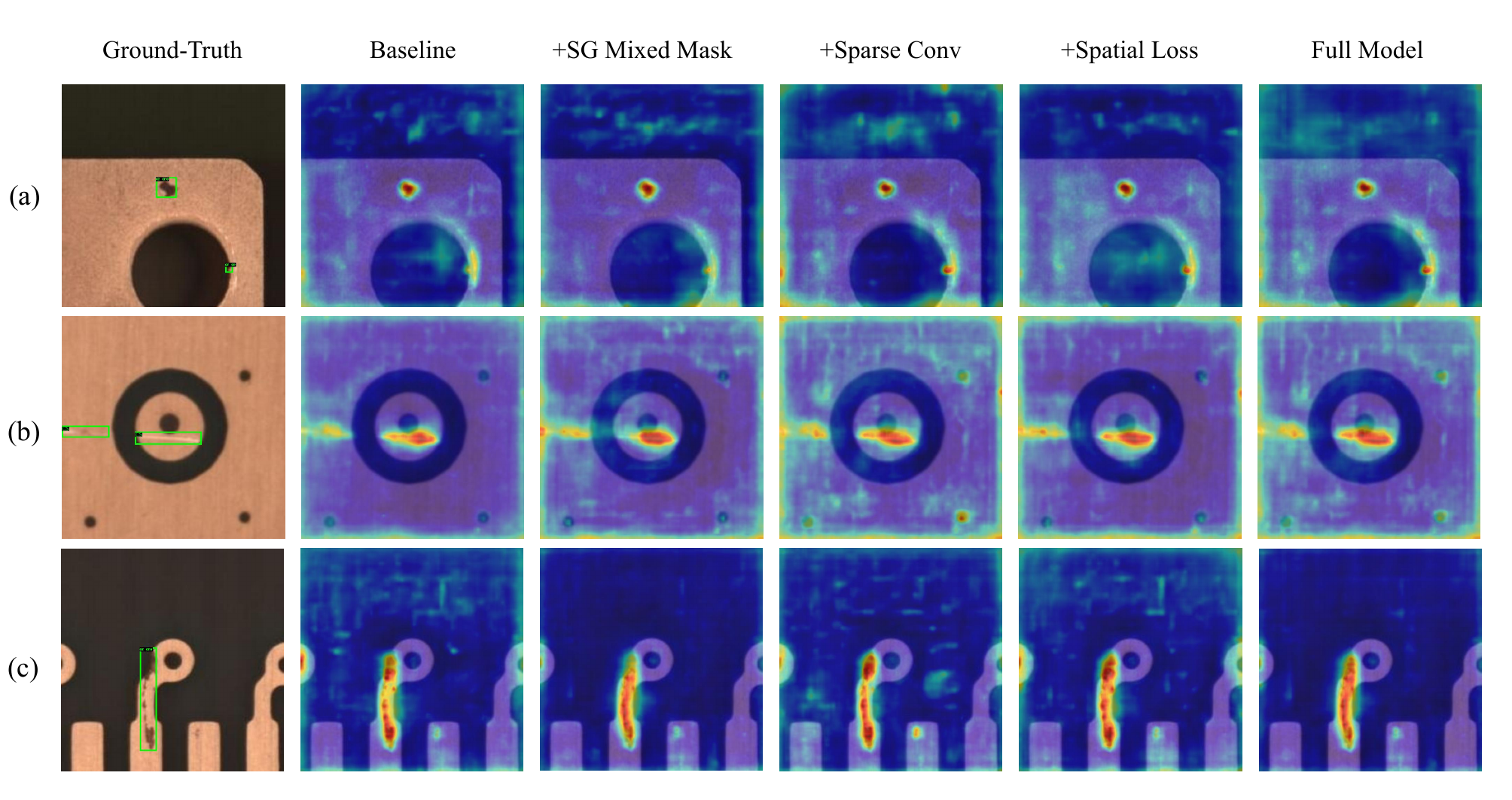}
    \caption{Qualitative visualization of output heat maps for different component settings in the ablation study.}
    \label{fig:ablation_vis}
\end{figure}

\subsection{Ablation Studies}

To comprehensively evaluate the proposed framework, we conduct ablation experiments on the DsPCBSD+ validation set from four aspects: the contribution of each module, the source of the pretraining gains, the refinement of key designs in Stage-A, and the refinement of key parameters in Stage-B. Through these experiments, we aim to verify not only whether each proposed module is effective, but also whether the overall design is reasonable and consistent from both quantitative and qualitative perspectives.

\subsubsection{Contribution of Each Module}
We conduct a component ablation study to evaluate the roles of the three proposed designs: SG mixed masking, sparse convolutional pretraining, and spatial continuity regularization. 
As shown in Table~\ref{tab:component_ablation}, all component combinations are tested under the same YOLOv8s detector setting. 
This comparison helps examine both the individual contribution of each module and its interaction with the other modules.

Table~\ref{tab:component_ablation} shows different effects of the three proposed modules. 
When used alone, Spatial Loss improves both Precision and Recall. 
Precision increases from 81.9\% to 82.4\%, and Recall increases from 79.3\% to 80.4\%. 
The two Stage-A components show a different trend when they are used separately. 
With SG Mixed Mask alone, Precision decreases from 81.9\% to 81.4\%, while Recall increases from 79.3\% to 80.1\%. 
With Sparse Conv. alone, Precision decreases from 81.9\% to 81.5\%, while Recall slightly increases from 79.3\% to 79.5\%. 
Each pretraining component alone makes the detector less conservative and helps it recover more potential defect instances, but the improvement in representation is still limited.
When SG Mixed Mask and Sparse Conv. are used together to form the SG-MIM pretraining stage, the behavior becomes more balanced. 
Precision increases to 82.2\%, Recall increases to 80.6\%, and $\mathrm{mAP}_{0.5:0.95}$ reaches 52.0\%. 
These results are higher than using either Stage-A component alone. 
The two Stage-A modules form an expected physical synergy rather than a simple stacking of components.

Upon incorporating the Spatial Loss into the SG-MIM pretrained detector, the complete model attains the highest overall performance. 
Relative to the baseline, the improvements amount to 1.4 and 1.6 percentage points, respectively. 
These findings indicate that each module contributes a distinct, non-redundant function within the proposed framework. 
When integrated, the modules exhibit a synergistic effect, collectively enhancing both detection accuracy and localization performance in a more coordinated manner.

Fig.~\ref{fig:ablation_vis} qualitatively supports the quantitative results. 
In case (a), the defect is small and weak. The baseline already highlights it locally but mixes it with background. SG mixed masking makes the defect response more distinguishable, Sparse Conv. suppresses irrelevant activation, and the full model keeps a compact response on the defect, consistent with Stage-A pretraining for weak-defect perception. 
In case (b), the defect is locally continuous. The baseline focuses on only part of it, while SG mixed masking and Sparse Conv. extend the activation more completely along the defect. This matches Stage-A’s goal: mixed masking encourages inferring missing PCB structure, and sparse encoding reduces interference from masked regions, aligning features with the full structural extent of the defect. 
In case (c), the defect is slender and elongated. The baseline response is scattered; adding Spatial Loss makes the activation more compact and better aligned with the defect, and the full model preserves this compact pattern while suppressing nearby irrelevant activations. This aligns with the purpose of spatial continuity regularization: reducing fragmented responses on continuous defect regions.

\begin{table}[t]
\centering
\caption{Comparison of different pretraining methods on the DsPCBSD+ validation set.}
\label{tab:pretrain_ablation}
\small
\setlength{\tabcolsep}{3.5pt}
\begin{tabular}{lcccc}
\hline
Pretraining & Precision & Recall & $\mathrm{mAP}_{0.5}$ (\%) & $\mathrm{mAP}_{0.5:0.95}$ (\%) \\
\hline
No pretraining & 81.9 & 79.3 & 84.1 & 50.7 \\
Official weights & 81.5{\scriptsize\textcolor{blue}{$\,\downarrow$0.4}} & 79.5{\scriptsize\textcolor{red}{$\,\uparrow$0.2}} & 84.2{\scriptsize\textcolor{red}{$\,\uparrow$0.1}} & 50.8{\scriptsize\textcolor{red}{$\,\uparrow$0.1}} \\
SimMIM~\cite{xie2022simmim} & \textbf{82.8}{\scriptsize\textcolor{red}{\textbf{$\,\uparrow$0.9}}} & 79.2{\scriptsize\textcolor{blue}{$\,\downarrow$0.1}} & 84.7{\scriptsize\textcolor{red}{$\,\uparrow$0.6}} & \underline{51.7}{\scriptsize\textcolor{red}{$\,\uparrow$1.0}} \\
MAE~\cite{he2022masked} & 80.8{\scriptsize\textcolor{blue}{$\,\downarrow$1.1}} & 80.1{\scriptsize\textcolor{red}{$\,\uparrow$0.8}} & 84.4{\scriptsize\textcolor{red}{$\,\uparrow$0.3}} & 50.9{\scriptsize\textcolor{red}{$\,\uparrow$0.2}} \\
BEiT~\cite{bao2021beit} & 80.7{\scriptsize\textcolor{blue}{$\,\downarrow$1.2}} & \textbf{80.9}{\scriptsize\textcolor{red}{\textbf{$\,\uparrow$1.6}}} & \underline{84.8}{\scriptsize\textcolor{red}{$\,\uparrow$0.7}} & \underline{51.7}{\scriptsize\textcolor{red}{$\,\uparrow$1.0}} \\
iBOT~\cite{zhou2022ibot} & 81.9{\scriptsize\textcolor{red}{$\,\uparrow$0.0}} & 79.5{\scriptsize\textcolor{red}{$\,\uparrow$0.2}} & 84.3{\scriptsize\textcolor{red}{$\,\uparrow$0.2}} & 51.4{\scriptsize\textcolor{red}{$\,\uparrow$0.7}} \\
\textbf{SG-MIM} & \underline{82.2}{\scriptsize\textcolor{red}{$\,\uparrow$0.3}} & \underline{80.6}{\scriptsize\textcolor{red}{$\,\uparrow$1.3}} & \textbf{85.2}{\scriptsize\textcolor{red}{\textbf{$\,\uparrow$1.1}}} & \textbf{52.0}{\scriptsize\textcolor{red}{\textbf{$\,\uparrow$1.3}}} \\
\hline
\end{tabular}
\end{table}

\subsubsection{Comparison of Different Pretraining Methods}

To further evaluate the effectiveness of the proposed SG-MIM pretraining strategy, we compare it with several representative pretraining methods under the same downstream detection setting. The goal of this experiment is to distinguish whether the observed gains mainly come from pretraining itself or from the design introduced for PCB defect detection in our method. Therefore, besides the baseline without pretraining, we include several widely used generic masked pretraining methods for comparison, and further provide heatmaps for qualitative analysis. To ensure a fair comparison, spatial continuity regularization is not introduced in this experiment, so that the differences among models can be attributed primarily to the pretraining strategy itself.
% 中文翻译：
% 为了进一步评估所提出的 SG-MIM 预训练策略的有效性，我们在相同的下游检测设置下，将其与若干具有代表性的预训练方法进行比较。
% 该实验的目的在于区分性能提升究竟主要来源于“使用了预训练”这一事实，还是来源于本文方法中引入的面向 PCB 的设计。
% 因此，除了不使用预训练的基线模型之外，我们还纳入了几种常用的通用掩码预训练方法进行对比，并进一步通过特征响应热力图开展定性分析。
% 为了保证比较公平，本组实验中不引入空间连续性正则化，从而使各模型之间的差异能够主要归因于预训练策略本身。

As reported in Table~\ref{tab:pretrain_ablation}, different pretraining methods lead to different downstream results. 
Compared with no pretraining, most methods improve $\mathrm{mAP}_{0.5}$ and $\mathrm{mAP}_{0.5:0.95}$, but the improvement is not equally stable across metrics. 
For example, MAE increases Recall, but only improves $\mathrm{mAP}_{0.5}$ from 84.1\% to 84.4\% and $\mathrm{mAP}_{0.5:0.95}$ from 50.7\% to 50.9\%. 
SimMIM performs better than MAE on the two mAP metrics, reaching 84.7\% and 51.7\%, respectively, but its Recall slightly decreases. 
In comparison, SG-MIM achieves 85.2\% in $\mathrm{mAP}_{0.5}$ and 52.0\% in $\mathrm{mAP}_{0.5:0.95}$, while also increasing Recall to 80.6\%. 
This shows that SG-MIM provides a better balance between defect coverage and localization quality than generic masked pretraining methods. 
The design introduced for PCB defect detection is more suitable for learning structural representations from PCB images.
\begin{figure}[t]
    \centering
    \includegraphics[width=1.0\textwidth]{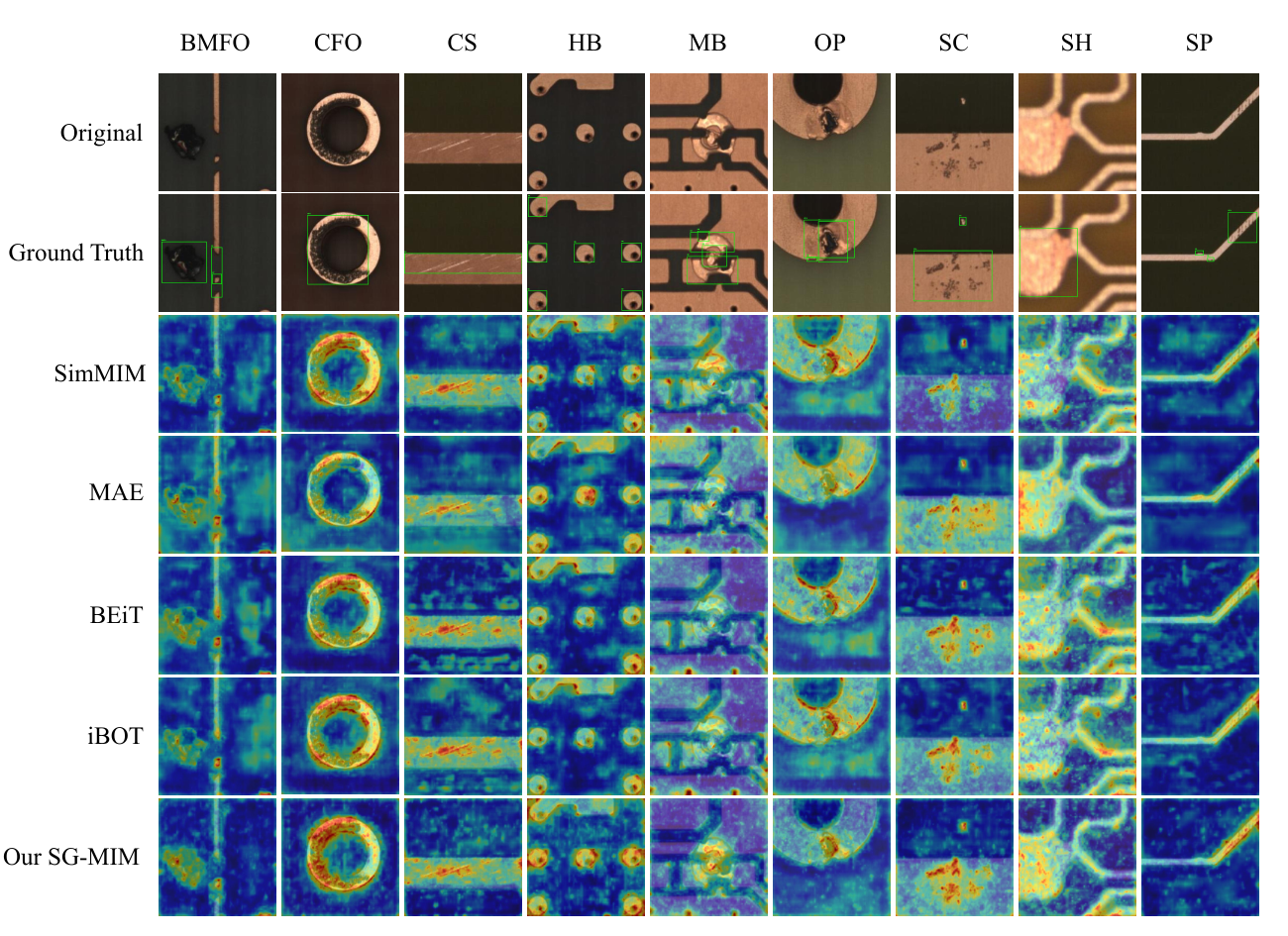}
    \caption{Visualization comparison of feature-response heatmaps produced by different pretrained feature extractors.}
    \label{fig:detection_compare}
\end{figure}

Fig.~\ref{fig:detection_compare} further presents the feature response heatmaps produced by different pretrained feature extractors on nine representative defect categories. 
Compared with representative pretraining methods, SG-MIM produces more focused responses around the main defect structures and reduces irrelevant background activation. 
This tendency is particularly evident for samples with weak contrast, tiny defect regions, or complex surrounding patterns, where generic pretraining methods often produce more diffuse or partially shifted responses. In contrast, SG-MIM highlights defect-relevant structures more clearly and preserves the underlying PCB layout more effectively, indicating that the learned representations are more compatible with the structural characteristics of PCB images. These qualitative observations are consistent with the quantitative improvements in Table~\ref{tab:pretrain_ablation} and further verify the effectiveness of the proposed structure-guided masked pretraining strategy.

\subsubsection{Effect of the Random and Structure-Guided Proportion}

To further evaluate the effect of the hybrid masking strategy in SG-MIM, we compare different sampling proportions between random masking and structure-guided masking. 
In this experiment, the complete SG-MIM pretraining framework is retained, including the same YOLOv8s backbone, sparse convolutional reconstruction pipeline, reconstruction objective, and training schedule, while spatial continuity regularization is still not introduced in Stage-B. 
Therefore, the performance differences in Table~\ref{tab:mask_proportions_comparison} can be mainly attributed to the sampling proportion itself.
% 中文翻译：
% 我们在相同的下游检测流程下比较了不同随机掩码与结构引导掩码比例的性能差异为了进一步评估 SG-MIM 中掩码策略的作用。在该实验中，我们保留了完整的 SG-MIM 预训练框架，包括相同的 YOLOv8 骨干网络、基于稀疏卷积的重建管线、重建目标以及训练设置，而在 Stage-B 中不引入空间连续性正则化。因此，表~\ref{tab:mask_ratio_comparison} 中的性能差异可以主要归因于掩码比例本身。

\begin{table}[t]
\centering
\caption{Comparison of different proportions between random masking and structure-guided masking.}
\label{tab:mask_proportions_comparison}
\normalsize
\setlength{\tabcolsep}{3.5pt}
\begin{tabular}{ccccc}
\hline
Rand.:SG & Precision & Recall & $\mathrm{mAP}_{0.5}$ (\%) & $\mathrm{mAP}_{0.5:0.95}$ (\%) \\
\hline
100 : 0 & 82.6 & 79.3 & 84.6 & 51.5 \\
70 : 30 & \underline{83.1}{\scriptsize\textcolor{red}{$\,\uparrow$0.5}} & 79.1{\scriptsize\textcolor{blue}{$\,\downarrow$0.2}} & 84.8{\scriptsize\textcolor{red}{$\,\uparrow$0.2}} & 51.6{\scriptsize\textcolor{red}{$\,\uparrow$0.1}} \\
50 : 50 & 81.9{\scriptsize\textcolor{blue}{$\,\downarrow$0.7}} & 79.0{\scriptsize\textcolor{blue}{$\,\downarrow$0.3}} & 84.9{\scriptsize\textcolor{red}{$\,\uparrow$0.3}} & 51.3{\scriptsize\textcolor{blue}{$\,\downarrow$0.2}} \\
40 : 60 & \textbf{83.5}{\scriptsize\textcolor{red}{\textbf{$\,\uparrow$0.9}}} & 79.7{\scriptsize\textcolor{red}{$\,\uparrow$0.4}} & \textbf{85.3}{\scriptsize\textcolor{red}{\textbf{$\,\uparrow$0.7}}} & 51.8{\scriptsize\textcolor{red}{$\,\uparrow$0.3}} \\
30 : 70 & 82.2{\scriptsize\textcolor{blue}{$\,\downarrow$0.4}} & \underline{80.6}{\scriptsize\textcolor{red}{$\,\uparrow$1.3}} & \underline{85.2}{\scriptsize\textcolor{red}{$\,\uparrow$0.6}} & \textbf{52.0}{\scriptsize\textcolor{red}{\textbf{$\,\uparrow$0.5}}} \\
20 : 80 & 81.3{\scriptsize\textcolor{blue}{$\,\downarrow$1.3}} & 79.8{\scriptsize\textcolor{red}{$\,\uparrow$0.5}} & \textbf{85.3}{\scriptsize\textcolor{red}{\textbf{$\,\uparrow$0.7}}} & 51.5{\scriptsize\textcolor{red}{$\,\uparrow$0.0}} \\
10 : 90 & 81.2{\scriptsize\textcolor{blue}{$\,\downarrow$1.4}} & 80.2{\scriptsize\textcolor{red}{$\,\uparrow$0.9}} & 85.0{\scriptsize\textcolor{red}{$\,\uparrow$0.4}} & \underline{51.9}{\scriptsize\textcolor{red}{$\,\uparrow$0.4}} \\
0 : 100 & 82.0{\scriptsize\textcolor{blue}{$\,\downarrow$0.6}} & \textbf{80.8}{\scriptsize\textcolor{red}{\textbf{$\,\uparrow$1.5}}} & 85.0{\scriptsize\textcolor{red}{$\,\uparrow$0.4}} & 51.8{\scriptsize\textcolor{red}{$\,\uparrow$0.3}} \\
\hline
\end{tabular}
\end{table}

Table~\ref{tab:mask_proportions_comparison} compares different sampling proportions between random masking and structure-guided masking in SG-MIM. 
Besides the two extreme settings, we test several mixed settings and use denser proportions in the structure-guided dominant range, such as $40{:}60$, $30{:}70$, $20{:}80$, and $10{:}90$. 
The results show that adding structure-guided masking generally improves the detection performance compared with the fully random setting. 
Among these settings, the best performance is obtained when the random-to-structure-guided proportion is set to $30{:}70$. 
However, the purely structure-guided setting does not achieve the best result. 
This is because it tends to mask most structurally important regions, which may make the reconstruction task overly concentrated on local high response structures and reduce the available contextual diversity. 
Therefore, the mixed strategy provides a better balance than either fully random masking or fully structure-guided masking.

\subsubsection{Effect of Spatial Continuity Regularization}

We further investigate the effect of the spatial continuity regularization coefficient $\lambda$ while fixing the SG-MIM pretrained initialization and all other settings unchanged. This experiment is intended to evaluate how different regularization strengths affect the detector after SG-MIM pretraining, and to identify a suitable value of $\lambda$ for the proposed spatial continuity constraint.
% 在固定 SG-MIM 预训练初始化及其他设置不变的条件下，
% 我们进一步研究空间连续性正则化系数 $\lambda$ 的影响。
% 该实验旨在评估在 SG-MIM 预训练之后，不同正则化强度对检测器性能的影响，
% 并为所提出的空间连续性约束确定一个合适的 $\lambda$ 取值。

\begin{table}[t]
\centering
\caption{Effect of the spatial continuity regularization coefficient $\lambda$.}
\label{tab:lambda_ablation}
\normalsize
\setlength{\tabcolsep}{6.5pt}
\begin{tabular}{ccccc}
\hline
$\lambda$ & Precision & Recall & $\mathrm{mAP}_{0.5}$ (\%) & $\mathrm{mAP}_{0.5:0.95}$ (\%) \\
\hline
0    & 82.2 & 80.6 & 85.2 & 52.0 \\
0.01 & 82.4{\scriptsize\textcolor{red}{$\,\uparrow$0.2}} & 80.4{\scriptsize\textcolor{blue}{$\,\downarrow$0.2}} & 85.0{\scriptsize\textcolor{blue}{$\,\downarrow$0.2}} & 51.6{\scriptsize\textcolor{blue}{$\,\downarrow$0.4}} \\
0.05 & \underline{83.4}{\scriptsize\textcolor{red}{$\,\uparrow$1.2}} & 78.5{\scriptsize\textcolor{blue}{$\,\downarrow$2.1}} & 85.2{\scriptsize\textcolor{red}{$\,\uparrow$0.0}} & 51.9{\scriptsize\textcolor{blue}{$\,\downarrow$0.1}} \\
0.10 & \textbf{83.5}{\scriptsize\textcolor{red}{\textbf{$\,\uparrow$1.3}}} & 79.4{\scriptsize\textcolor{blue}{$\,\downarrow$1.2}} & 85.4{\scriptsize\textcolor{red}{$\,\uparrow$0.2}} & 51.9{\scriptsize\textcolor{blue}{$\,\downarrow$0.1}} \\
0.15 & 82.7{\scriptsize\textcolor{red}{$\,\uparrow$0.5}} & 79.8{\scriptsize\textcolor{blue}{$\,\downarrow$0.8}} & 85.3{\scriptsize\textcolor{red}{$\,\uparrow$0.1}} & 52.1{\scriptsize\textcolor{red}{$\,\uparrow$0.1}} \\
0.20 & 81.6{\scriptsize\textcolor{blue}{$\,\downarrow$0.6}} & \underline{80.8}{\scriptsize\textcolor{red}{$\,\uparrow$0.2}} & \underline{85.5}{\scriptsize\textcolor{red}{$\,\uparrow$0.3}} & \textbf{52.3}{\scriptsize\textcolor{red}{\textbf{$\,\uparrow$0.3}}} \\
0.25 & 81.1{\scriptsize\textcolor{blue}{$\,\downarrow$1.1}} & 80.6{\scriptsize\textcolor{red}{$\,\uparrow$0.0}} & 84.4{\scriptsize\textcolor{blue}{$\,\downarrow$0.8}} & 51.7{\scriptsize\textcolor{blue}{$\,\downarrow$0.3}} \\
0.30 & 81.7{\scriptsize\textcolor{blue}{$\,\downarrow$0.5}} & \textbf{81.2}{\scriptsize\textcolor{red}{\textbf{$\,\uparrow$0.6}}} & \textbf{85.7}{\scriptsize\textcolor{red}{\textbf{$\,\uparrow$0.5}}} & \underline{52.2}{\scriptsize\textcolor{red}{$\,\uparrow$0.2}} \\
\hline
\end{tabular}
\end{table}

From Table~\ref{tab:lambda_ablation}, it can be observed that the effect of spatial continuity regularization depends on the choice of $\lambda$. Very small coefficients, such as $\lambda=0.01$, do not provide performance gains and even lead to slight decreases in recall and mAP, indicating that the regularization is too weak to produce a meaningful spatial constraint. As $\lambda$ increases to a moderate range, the detector begins to benefit more clearly from the proposed regularization. In particular, Precision reaches its highest value at $\lambda=0.10$, while Recall continues to improve and attains the best result at $\lambda=0.30$. Meanwhile, $\mathrm{mAP}_{0.5}$ achieves its best value at $\lambda=0.30$, whereas $\mathrm{mAP}_{0.5:0.95}$ reaches its maximum at $\lambda=0.20$. The proposed spatial continuity term mainly improves prediction consistency and localization quality during fine-tuning, especially under moderate regularization strength. Considering both overall detection accuracy and stability, $\lambda=0.20$ provides the most balanced performance and is therefore adopted as the default setting in our method.

\section{Conclusion}
\label{sec:conclusion}

This paper presented a two-phase PCB defect detection framework that combines SG-MIM pretraining with spatial continuity regularization in YOLOv8. 
In Stage-A, the backbone learns PCB structural priors from unlabeled images through structure-guided mixed masking and sparse convolutional masked pretraining. 
In Stage-B, spatial continuity regularization improves localization coherence by reducing scattered same-class predictions during fine-tuning. 
Experiments on the DsPCBSD+ dataset show that the proposed framework improves both representation learning and detection performance.

Although the current pretraining stage only uses unlabeled images from the existing dataset, it already brings clear gains. 
This demonstrates that SG-MIM may benefit further from larger and more diverse unlabeled PCB images. 
In future work, we will explore more adaptive structure prior extraction, stronger continuity constraints for irregular defects, and broader validation on other industrial AOI datasets. 
We will also investigate lighter implementations to improve deployment efficiency in real production scenarios.

% 本文提出了一种结合SG-MIM预训练与空间连续性正则化的两阶段PCB缺陷检测框架，基于YOLOv8实现。在阶段A，骨干网络通过结构引导混合掩码与稀疏卷积掩码预训练，从无标注图像中学习PCB结构先验。在阶段B，空间连续性正则化通过减少微调过程中同类预测的分散分布，提升了定位的一致性。在DsPCBSD+数据集上的实验表明，所提框架同时改善了表征学习与检测性能。

% 尽管当前预训练阶段仅使用了现有数据集中的无标注图像，但仍带来了明显的性能增益。这表明SG-MIM未来有望从规模更大、更多样的无标注PCB图像中获得进一步收益。在后续工作中，我们将探索更自适应的结构先验提取方法、针对不规则缺陷的更强连续性约束，以及其他工业自动光学检测数据集上的更广泛验证。同时，我们将研究更轻量化的实现方案，以提升实际生产场景中的部署效率。

\section*{Data Availability}

The DsPCBSD+ dataset used in this study is publicly available from the source cited in this paper~\cite{lv2024dataset}. The data split and experimental configuration files will be made available with the source code upon acceptance.

\section*{Code Availability}

The source code will be made publicly available upon acceptance.

% % =========================================================
% \section*{CRediT authorship contribution statement}
% \textbf{First Author:} Conceptualization, Methodology, Software, Writing -- original draft.
% \textbf{Second Author:} Methodology, Software, Writing -- review \& editing.
% \textbf{Last Author:} Supervision, Validation, Writing -- review \& editing.

% \section*{Declaration of Competing Interest}
% The authors declare that they have no known competing financial interests or personal relationships that could have appeared to influence the work reported in this paper.

% \section*{Acknowledgements}
% This work was supported by XXX (if applicable). The authors thank XXX for helpful discussions.

% \section*{Data availability}
% Data will be made available on request (or provide a link if you plan to release it).

\bibliographystyle{elsarticle-num}
\bibliography{cas-refs}

\begin{thebibliography}{10}
\expandafter\ifx\csname url\endcsname\relax
  \def\url#1{\texttt{#1}}\fi
\expandafter\ifx\csname urlprefix\endcsname\relax\def\urlprefix{URL }\fi
\expandafter\ifx\csname href\endcsname\relax
  \def\href#1#2{#2} \def\path#1{#1}\fi

\bibitem{tang2024lightweight}
J.~Tang, Z.~Wang, H.~Zhang, H.~Li, P.~Wu, N.~Zeng, A lightweight surface defect detection framework combined with dual-domain attention mechanism, Expert Systems with Applications 238 (2024) 121726.

\bibitem{angelopoulos2019tackling}
A.~Angelopoulos, E.~T. Michailidis, N.~Nomikos, P.~Trakadas, A.~Hatziefremidis, S.~Voliotis, T.~Zahariadis, Tackling faults in the industry 4.0 era—a survey of machine-learning solutions and key aspects, Sensors 20~(1) (2019) 109.

\bibitem{TAN2020121892}
Q.~Tan, L.~Liu, M.~Yu, J.~Li, An innovative method of recycling metals in printed circuit board (pcb) using solutions from pcb production, Journal of Hazardous Materials 390 (2020) 121892.

\bibitem{MOGANTI1996287}
M.~Moganti, F.~Ercal, C.~H. Dagli, S.~Tsunekawa, Automatic pcb inspection algorithms: A survey, Computer Vision and Image Understanding 63~(2) (1996) 287--313.

\bibitem{ZHOU2023557}
Y.~Zhou, M.~Yuan, J.~Zhang, G.~Ding, S.~Qin, Review of vision-based defect detection research and its perspectives for printed circuit board, Journal of Manufacturing Systems 70 (2023) 557--578.

\bibitem{10044670}
Q.~Ling, N.~A.~M. Isa, Printed circuit board defect detection methods based on image processing, machine learning and deep learning: A survey, IEEE Access 11 (2023) 15921--15944.

\bibitem{KANG2023120121}
D.~Kang, J.~Lai, Y.~Han, Improving surface defect detection with context-guided asymmetric modulation networks and confidence-boosting loss, Expert Systems with Applications 225 (2023) 120121.

\bibitem{SUN2025128101}
P.~Sun, C.~Hua, W.~Ding, C.~Hua, A real–time detection framework for surface defects in ceramic tableware based on deep learning, Expert Systems with Applications 286 (2025) 128101.

\bibitem{MENG2025113578}
S.~Meng, S.~Zhang, X.~Liang, J.~Hu, Automatic extraction of scale information for interactive measurement of anything in microscopy images, Knowledge-Based Systems 324 (2025) 113578.

\bibitem{7485869}
S.~Ren, K.~He, R.~Girshick, J.~Sun, Faster r-cnn: Towards real-time object detection with region proposal networks, IEEE Transactions on Pattern Analysis and Machine Intelligence 39~(6) (2017) 1137--1149.

\bibitem{10129965}
C.~Song, J.~Chen, Z.~Lu, F.~Li, Y.~Liu, Steel surface defect detection via deformable convolution and background suppression, IEEE Transactions on Instrumentation and Measurement 72 (2023) 1--9.

\bibitem{liu2016ssd}
W.~Liu, D.~Anguelov, D.~Erhan, C.~Szegedy, S.~Reed, C.-Y. Fu, A.~C. Berg, Ssd: Single shot multibox detector, in: European conference on computer vision, Springer, 2016, pp. 21--37.

\bibitem{Duan_2019_ICCV}
K.~Duan, S.~Bai, L.~Xie, H.~Qi, Q.~Huang, Q.~Tian, Centernet: Keypoint triplets for object detection, in: Proceedings of the IEEE/CVF International Conference on Computer Vision (ICCV), 2019.

\bibitem{dosovitskiy2020image}
A.~Dosovitskiy, L.~Beyer, A.~Kolesnikov, D.~Weissenborn, X.~Zhai, T.~Unterthiner, M.~Dehghani, M.~Minderer, G.~Heigold, S.~Gelly, et~al., An image is worth 16x16 words: Transformers for image recognition at scale, arXiv preprint arXiv:2010.11929 (2020).

\bibitem{Liu_2021_ICCV}
Z.~Liu, Y.~Lin, Y.~Cao, H.~Hu, Y.~Wei, Z.~Zhang, S.~Lin, B.~Guo, Swin transformer: Hierarchical vision transformer using shifted windows, in: Proceedings of the IEEE/CVF International Conference on Computer Vision (ICCV), 2021, pp. 10012--10022.

\bibitem{carion2020end}
N.~Carion, F.~Massa, G.~Synnaeve, N.~Usunier, A.~Kirillov, S.~Zagoruyko, End-to-end object detection with transformers, in: European conference on computer vision, Springer, 2020, pp. 213--229.

\bibitem{zong2023codetr}
Z.~Zong, G.~Song, Y.~Liu, Detrs with collaborative hybrid assignments training, in: Proceedings of the IEEE/CVF International Conference on Computer Vision (ICCV), 2023, pp. 19609--19619.

\bibitem{ma2024surface}
Y.~Ma, J.~Yin, F.~Huang, Q.~Li, Surface defect inspection of industrial products with object detection deep networks: A systematic review, Artificial Intelligence Review 57~(12) (2024) 333.

\bibitem{zhu2025novel}
L.~Zhu, R.~Zhao, A novel pcb surface defect detection method based on separated global context attention to guide residual context aggregation, Scientific Reports 15~(1) (2025) 9620.

\bibitem{khan2023survey}
A.~Khan, Z.~Rauf, A.~Sohail, A.~R. Khan, H.~Asif, A.~Asif, U.~Farooq, A survey of the vision transformers and their cnn-transformer based variants, Artificial Intelligence Review 56~(Suppl 3) (2023) 2917--2970.

\bibitem{rs16162910}
Q.~Yuan, Y.~Shi, M.~Li, A review of computer vision-based crack detection methods in civil infrastructure: Progress and challenges, Remote Sensing 16~(16) (2024).

\bibitem{app14156774}
Y.~He, S.~Li, X.~Wen, J.~Xu, A survey on surface defect inspection based on generative models in manufacturing, Applied Sciences 14~(15) (2024).

\bibitem{sohan2024review}
M.~Sohan, T.~Sai~Ram, C.~V. Rami~Reddy, A review on yolov8 and its advancements, in: International conference on data intelligence and cognitive informatics, Springer, 2024, pp. 529--545.

\bibitem{yaseen2024yolov8indepthexplorationinternal}
M.~Yaseen, What is yolov8: An in-depth exploration of the internal features of the next-generation object detector, arXiv preprint arXiv:2408.15857 (2024).

\bibitem{Zhao2024}
Q.~Zhao, T.~Ji, S.~Liang, W.~Yu, Pcb surface defect fast detection method based on attention and multi-source fusion, Multimedia Tools and Applications 83~(2) (2024) 5451--5472.

\bibitem{10.1117/1.JEI.30.4.043004}
G.~Liu, H.~Wen, {Printed circuit board defect detection based on MobileNet-Yolo-Fast}, Journal of Electronic Imaging 30~(4) (2021) 043004.

\bibitem{tang2023pcb}
J.~Tang, S.~Liu, D.~Zhao, L.~Tang, W.~Zou, B.~Zheng, Pcb-yolo: An improved detection algorithm of pcb surface defects based on yolov5, Sustainability 15~(7) (2023) 5963.

\bibitem{9905499}
W.~Xuan, G.~Jian-She, H.~Bo-Jie, W.~Zong-Shan, D.~Hong-Wei, W.~Jie, A lightweight modified yolox network using coordinate attention mechanism for pcb surface defect detection, IEEE Sensors Journal 22~(21) (2022) 20910--20920.

\bibitem{LIU2022116178}
X.~Liu, J.~Hu, H.~Wang, Z.~Zhang, X.~Lu, C.~Sheng, S.~Song, J.~Nie, Gaussian-iou loss: Better learning for bounding box regression on pcb component detection, Expert Systems with Applications 190 (2022) 116178.

\bibitem{yuan2024yolo}
M.~Yuan, Y.~Zhou, X.~Ren, H.~Zhi, J.~Zhang, H.~Chen, Yolo-hmc: An improved method for pcb surface defect detection, IEEE Transactions on Instrumentation and Measurement 73 (2024) 1--11.

\bibitem{ling2023printed}
Q.~Ling, N.~A.~M. Isa, Printed circuit board defect detection methods based on image processing, machine learning and deep learning: A survey, IEEE access 11 (2023) 15921--15944.

\bibitem{zhou2023review}
Y.~Zhou, M.~Yuan, J.~Zhang, G.~Ding, S.~Qin, Review of vision-based defect detection research and its perspectives for printed circuit board, Journal of Manufacturing Systems 70 (2023) 557--578.

\bibitem{tao2022deep}
X.~Tao, X.~Gong, X.~Zhang, S.~Yan, C.~Adak, Deep learning for unsupervised anomaly localization in industrial images: A survey, IEEE Transactions on Instrumentation and Measurement 71 (2022) 1--21.

\bibitem{jing2020self}
L.~Jing, Y.~Tian, Self-supervised visual feature learning with deep neural networks: A survey, IEEE transactions on pattern analysis and machine intelligence 43~(11) (2020) 4037--4058.

\bibitem{oord2018representation}
A.~v.~d. Oord, Y.~Li, O.~Vinyals, Representation learning with contrastive predictive coding, arXiv preprint arXiv:1807.03748 (2018).

\bibitem{he2020momentum}
K.~He, H.~Fan, Y.~Wu, S.~Xie, R.~Girshick, Momentum contrast for unsupervised visual representation learning, in: Proceedings of the IEEE/CVF conference on computer vision and pattern recognition, 2020, pp. 9729--9738.

\bibitem{chen2020simple}
T.~Chen, S.~Kornblith, M.~Norouzi, G.~Hinton, A simple framework for contrastive learning of visual representations, in: International conference on machine learning, PmLR, 2020, pp. 1597--1607.

\bibitem{grill2020bootstrap}
J.-B. Grill, F.~Strub, F.~Altch{\'e}, C.~Tallec, P.~Richemond, E.~Buchatskaya, C.~Doersch, B.~Avila~Pires, Z.~Guo, M.~Gheshlaghi~Azar, et~al., Bootstrap your own latent-a new approach to self-supervised learning, Advances in neural information processing systems 33 (2020) 21271--21284.

\bibitem{caron2020unsupervised}
M.~Caron, I.~Misra, J.~Mairal, P.~Goyal, P.~Bojanowski, A.~Joulin, Unsupervised learning of visual features by contrasting cluster assignments, Advances in neural information processing systems 33 (2020) 9912--9924.

\bibitem{chen2021exploring}
X.~Chen, K.~He, Exploring simple siamese representation learning, in: Proceedings of the IEEE/CVF conference on computer vision and pattern recognition, 2021, pp. 15750--15758.

\bibitem{vaswani2017attention}
A.~Vaswani, N.~Shazeer, N.~Parmar, J.~Uszkoreit, L.~Jones, A.~N. Gomez, {\L}.~Kaiser, I.~Polosukhin, Attention is all you need, Advances in neural information processing systems 30 (2017).

\bibitem{devlin2019bert}
J.~Devlin, M.-W. Chang, K.~Lee, K.~Toutanova, Bert: Pre-training of deep bidirectional transformers for language understanding, in: Proceedings of the 2019 conference of the North American chapter of the association for computational linguistics: human language technologies, volume 1 (long and short papers), 2019, pp. 4171--4186.

\bibitem{bao2021beit}
H.~Bao, L.~Dong, S.~Piao, F.~Wei, Beit: Bert pre-training of image transformers, arXiv preprint arXiv:2106.08254 (2021).

\bibitem{he2022masked}
K.~He, X.~Chen, S.~Xie, Y.~Li, P.~Doll{\'a}r, R.~Girshick, Masked autoencoders are scalable vision learners, in: Proceedings of the IEEE/CVF conference on computer vision and pattern recognition, 2022, pp. 16000--16009.

\bibitem{xie2022simmim}
Z.~Xie, Z.~Zhang, Y.~Cao, Y.~Lin, J.~Bao, Z.~Yao, Q.~Dai, H.~Hu, Simmim: A simple framework for masked image modeling, in: Proceedings of the IEEE/CVF conference on computer vision and pattern recognition, 2022, pp. 9653--9663.

\bibitem{zhou2023self}
L.~Zhou, H.~Liu, J.~Bae, J.~He, D.~Samaras, P.~Prasanna, Self pre-training with masked autoencoders for medical image classification and segmentation, in: 2023 IEEE 20th international symposium on biomedical imaging (ISBI), IEEE, 2023, pp. 1--6.

\bibitem{hondru2025masked}
V.~Hondru, F.~A. Croitoru, S.~Minaee, R.~T. Ionescu, N.~Sebe, Masked image modeling: A survey, International Journal of Computer Vision 133~(10) (2025) 7154--7200.

\bibitem{tian2023designing}
K.~Tian, Y.~Jiang, Q.~Diao, C.~Lin, L.~Wang, Z.~Yuan, Designing bert for convolutional networks: Sparse and hierarchical masked modeling, arXiv preprint arXiv:2301.03580 (2023).

\bibitem{canny2009computational}
J.~Canny, A computational approach to edge detection, IEEE Transactions on Pattern Analysis and Machine Intelligence PAMI-8~(6) (1986) 679--698.

\bibitem{marr1980theory}
D.~Marr, E.~Hildreth, Theory of edge detection, Proceedings of the Royal Society of London. Series B. Biological Sciences 207~(1167) (1980) 187--217.

\bibitem{haralick2007textural}
R.~M. Haralick, K.~Shanmugam, I.~Dinstein, Textural features for image classification, IEEE Transactions on Systems, Man, and Cybernetics SMC-3~(6) (1973) 610--621.

\bibitem{freeman1991design}
W.~T. Freeman, E.~H. Adelson, et~al., The design and use of steerable filters, IEEE Transactions on Pattern analysis and machine intelligence 13~(9) (1991) 891--906.

\bibitem{bigun2002multidimensional}
J.~Bigun, G.~H. Granlund, J.~Wiklund, Multidimensional orientation estimation with applications to texture analysis and optical flow, IEEE Transactions on pattern analysis and machine intelligence 13~(8) (2002) 775--790.

\bibitem{lv2024dataset}
S.~Lv, B.~Ouyang, Z.~Deng, T.~Liang, S.~Jiang, K.~Zhang, J.~Chen, Z.~Li, A dataset for deep learning based detection of printed circuit board surface defect, Scientific Data 11~(1) (2024) 811.

\bibitem{chen2021parallel}
P.-Y. Chen, M.-C. Chang, J.-W. Hsieh, Y.-S. Chen, Parallel residual bi-fusion feature pyramid network for accurate single-shot object detection, IEEE transactions on Image Processing 30 (2021) 9099--9111.

\bibitem{xu2022pp}
S.~Xu, X.~Wang, W.~Lv, Q.~Chang, C.~Cui, K.~Deng, G.~Wang, Q.~Dang, S.~Wei, Y.~Du, et~al., Pp-yoloe: An evolved version of yolo, arXiv preprint arXiv:2203.16250 (2022).

\bibitem{yolov5}
G.~Jocher, Ultralytics yolov5, gitHub repository (2020).

\bibitem{xu2022damo}
X.~Xu, Y.~Jiang, W.~Chen, Y.~Huang, Y.~Zhang, X.~Sun, Damo-yolo: A report on real-time object detection design, arXiv preprint arXiv:2211.15444 (2022).

\bibitem{lyu2022rtmdet}
C.~Lyu, W.~Zhang, H.~Huang, Y.~Zhou, Y.~Wang, Y.~Liu, S.~Zhang, K.~Chen, Rtmdet: An empirical study of designing real-time object detectors, arXiv preprint arXiv:2212.07784 (2022).

\bibitem{zhao2024rtdetr}
Y.~Zhao, W.~Lv, S.~Xu, J.~Wei, G.~Wang, Q.~Dang, Y.~Liu, J.~Chen, Detrs beat yolos on real-time object detection, in: Proceedings of the IEEE/CVF Conference on Computer Vision and Pattern Recognition (CVPR), 2024, pp. 16965--16974.

\bibitem{li2023efd}
S.~Li, F.~Kong, R.~Wang, T.~Luo, Z.~Shi, Efd-yolov4: A steel surface defect detection network with encoder-decoder residual block and feature alignment module, Measurement 220 (2023) 113359.

\bibitem{ultralytics2023yolov8docs}
{Ultralytics}, Yolov8 documentation, Ultralytics official documentation (2023).

\bibitem{ultralytics2023yolo}
{Ultralytics}, Ultralytics yolo, GitHub repository (2023).

\bibitem{wang2024yolov10}
A.~Wang, H.~Chen, L.~Liu, K.~Chen, Z.~Lin, J.~Han, G.~Ding, Yolov10: Real-time end-to-end object detection, arXiv preprint arXiv:2405.14458 (2024).

\bibitem{li2022yolov6}
C.~Li, L.~Li, H.~Jiang, K.~Weng, Y.~Geng, L.~Li, Z.~Ke, Q.~Li, M.~Cheng, W.~Nie, Y.~Li, B.~Zhang, Y.~Liang, L.~Zhou, X.~Xu, X.~Chu, X.~Wei, X.~Wei, Yolov6: A single-stage object detection framework for industrial applications, arXiv preprint arXiv:2209.02976 (2022).

\bibitem{zhou2022ibot}
J.~Zhou, C.~Wei, H.~Wang, W.~Shen, C.~Xie, A.~Yuille, T.~Kong, ibot: Image bert pre-training with online tokenizer, arXiv preprint arXiv:2111.07832 (2021).

\end{thebibliography}
\end{document}